\documentclass[10pt,twocolumn]{article}

\usepackage[a4paper,margin=0.75in,columnsep=0.25in]{geometry}

\usepackage[utf8]{inputenc}
\usepackage[T1]{fontenc}
\usepackage{times}
\usepackage{microtype}
\usepackage{booktabs}
\usepackage{graphicx}
\usepackage{amsmath,amssymb}
\usepackage{hyperref}
\usepackage{xcolor}
\usepackage[small,compact]{titlesec}
\usepackage{enumitem}
\setlist{nosep,leftmargin=*}
\usepackage[font=small,labelfont=bf]{caption}
\usepackage{subcaption}
\usepackage{dsfont}
\usepackage{listings}
\usepackage[round,authoryear]{natbib}
\titlespacing*{\section}{0pt}{8pt}{4pt}
\titlespacing*{\subsection}{0pt}{6pt}{3pt}
\titlespacing*{\paragraph}{0pt}{4pt}{6pt}

\title{\textbf{Can AI Scientist Agents Learn from Lab-in-the-Loop Feedback?\\Evidence from Iterative Perturbation Discovery}}

\author{
  Gilles Wainrib,\quad
  Barbara Bodinier,\quad
  Haitem Dakhli,\quad
  Josep Monserrat,\\
  Almudena Espin Perez,\quad
  Sabrina Carpentier,\quad
  Roberta Codato,\quad 
  John Klein\\[4pt]
  \normalsize Owkin Inc.
}
\date{}

\begin{document}
\maketitle

\begin{abstract}
\noindent
Recent work has questioned whether large language models (LLMs) can perform genuine in-context learning (ICL) for scientific experimental design, with prior studies suggesting that LLM-based agents exhibit no sensitivity to experimental feedback. We shed new light on this question by carrying out 800 independently replicated experiments on iterative perturbation discovery in Cell Painting high-content screening. 
We compare an LLM agent that iteratively updates its hypotheses using experimental feedback to a zero-shot baseline that relies solely on pretraining knowledge retrieval. Access to feedback yields a +53.4\% increase in discoveries per feature on average ($p = 0.003$). To test whether this improvement arises from genuine feedback-driven learning rather than prompt-induced recall of pretraining knowledge, we introduce a random feedback control in which hit/miss labels are permuted. Under this control, the performance gain disappears, indicating that the observed improvement depends on the structure of the feedback signal ($+13.0$ hits, $p = 0.003$). We further examine how model capability affects feedback utilization. Upgrading from Claude Sonnet~4.5 to~4.6 reduces gene hallucination rates from ${\sim}33\%$--$45\%$ to ${\sim}3$--$9\%$, converting a non-significant ICL effect ($+0.8$, $p = 0.32$) into a large and highly significant improvement ($+11.0$, $p=0.003$) for the best ICL strategy. These results suggest that effective in-context learning from experimental feedback emerges only once models reach a sufficient capability threshold.\end{abstract}

\section{Introduction}

Large language models encode substantial biological knowledge from their training corpora, raising the prospect of LLM-powered agents that can guide scientific experimentation. BioDiscoveryAgent \citep{roohani2024biodiscoveryagent} demonstrated that LLM agents can design genetic perturbation experiments, achieving improvements over Bayesian optimization baselines. However, a fundamental question remains contested: do these agents actually \textit{learn} from experimental feedback, or do they merely leverage static prior knowledge?

\citet{gupta2025llms} present striking evidence supporting the latter view. They compare BioDiscoveryAgent with a variant receiving randomly permuted labels, finding comparable performance. They conclude that ``LLMs trained on next-token prediction and RLHF fail to perform in-context experimental design'' and propose LLMNN, restricting LLMs to prior-based selection while delegating iterative updates to classical nearest-neighbor methods.

We revisit this question with a large-scale empirical study. Using the JUMP Cell Painting dataset, the largest public morphological profiling resource \citep{chandrasekaran2023jump}, we conduct a large-scale benchmark comparing 6 agent architectures across 10 target features with 10 replicates each (800 total experiments). Our best feedback-enabled LLM agent achieves $+185\%$ improvement over random selection, compared to $+85\%$ for prior knowledge alone ($p = 0.003$) across features. The ICL effect is also significant within features on all 10 target features ($p < 0.01$), with gains ranging from $+3.7$ (F90) to $+27.4$ hits (F80) additional discoveries.

While the magnitude varies across features, the direction is consistently positive. We further show that model capability matters: upgrading from Claude Sonnet 4.5 to 4.6 boosts dramatically the ICL effect and reduced gene hallucination from ${\sim}33\%$--$45\%$ to ${\sim}3$--$9\%$, enabling the agent to translate its reasoning into effective experimental selections.

\section{Related Work}

\paragraph{ICL for closed-loop optimization.}
The application of LLMs within sequential optimization loops has accelerated rapidly. BioDiscoveryAgent \citep{roohani2024biodiscoveryagent} demonstrated 21\% improvement over Bayesian Optimization (BO) baselines for genetic perturbation design. Concurrent work applies similar approaches to chemical synthesis \citep{boiko2023autonomous,zhang2025large}, materials discovery \citep{abhyankar2025accelerating}, and drug design \citep{wang2025polo}. LLM-guided Bayesian optimization \citep{ramos2023bayesian,liu2024large} uses ICL as an acquisition function, replacing hand-crafted kernels with learned priors. \citet{chen2023evoprompting} demonstrate closed-loop evolutionary optimization where LLM-generated candidates are evaluated and high-performing solutions serve as in-context exemplars for subsequent generations.

\paragraph{Skepticism about ICL for experimental design.}
\citet{gupta2025llms} challenge whether LLMs genuinely learn from feedback. Their key experiment, comparing BioDiscoveryAgent with a variant receiving randomly permuted outcomes, finds comparable performance, suggesting improvements stem from prior knowledge rather than in-context adaptation. \citet{falck2024context} show that ICL violates exchangeability assumptions when context becomes sufficiently long. \citet{wang2024can} report limited out-of-distribution ICL generalization. These results collectively question whether ICL-based experimental design is a genuine capability or an artifact of encoded priors.

\paragraph{Cell Painting and morphological profiling.}
Cell Painting is a high-content imaging assay capturing cellular phenotypes through six fluorescent stains targeting distinct organelles \citep{bray2016cell}. The JUMP consortium has generated profiles for over 116,000 chemical and 22,000 genetic perturbations \citep{chandrasekaran2023jump}. \citet{lu2025cellclip} introduce CellCLIP to align language and Cell Painting representations, but do not address sequential perturbation selection. To our knowledge, no prior work uses ICL-based perturbation optimization on Cell Painting data. 

\section{Methods}

\subsection{Data and Task}

We use the CRISPR perturbation subset of the JUMP Cell Painting dataset, comprising ${\sim}8{,}000$ gene knockouts with morphological profiles across 4,672 CellProfiler features. For each target feature $y$, the hit-or-miss reward for a perturbation $x$ is $r_y(x) = \mathds{1}[p_y(x) < 0.05]$, where $p_y(x)$ is the reported p-value of knockout $x$ on feature $y$. Each experimental campaign consists of $T{=}10$ iterations with batch size $K{=}100$, totaling 1,000 perturbations from ${\sim}8{,}000$ available genes. Performance is measured by cumulative unique discoveries.

We evaluate on 10 target features spanning the feature difficulty spectrum (baseline hit rates ${\sim}1$--$2\%$). Each condition is evaluated with 10 independent replicates, for 100 total campaigns per agent type (800 experiments across 8 conditions), providing sufficient statistical power for formal hypothesis testing.

\subsection{Agent Architectures}

We compare three LLM agent variants, isolating the effect of experimental feedback. Our results use Claude Sonnet~4.6 (\texttt{claude-sonnet-4-6}) \citep{anthropic2025claude} and Sonnet~4.5 (\texttt{claude-sonnet-4-5-20250929}) as backbone LLM to study the effect of model capability:

\paragraph{Zero-shot (Prior Knowledge Only).}
The agent receives only the target feature description and list of available genes. It selects perturbations based solely on encoded biological knowledge. This agent receives \emph{no information} about previous experimental outcomes, providing a clean measure of prior knowledge contribution.

\paragraph{In-Context Learning from Experimental Feedback (ICL-EF).}
The agent observes all previous experimental results: which genes were tested and whether they produced significant effects by exposing p-values. The prompt includes the full outcome history, enabling the agent to identify patterns. For example, ``ATP6V1A was a hit, suggesting other V-ATPase subunits may also succeed.'' 
Additionally, to guide exploitation, the most frequent alphabetic prefixes among successful genes are appended to the prompt, see Appendix \ref{app:icl} for full details. 

\paragraph{In-Context Belief Revision from Experimental Feedback (ICBR-EF).}
This variant extends the ICL-EF approach by incorporating a phenotypic signature (the top 10 most significantly perturbed features alongside the target feature) into the prompting context and a structured way to accumulate evidence. Phenotypic signatures allow the agent to reason not just about binary success, but about the specific morphological phenotypes induced by each perturbation, mimicking a Bayesian update of underlying biological mechanisms. 
To do this, the agent maintains an explicit \textit{hypothesis register}, i.e. beliefs about which biological mechanisms underlie the target phenotype.
Each hypothesis is tracked using the following structure: a confidence level (High/Medium/Low) and a status (Active/Weakened/Abandoned).
At each iteration, the current register is serialized into the LLM prompt. The LLM returns a JSON object encapsulating the \textit{hypothesis register}, which replaces the current state. 
The register is entirely LLM-managed: the model itself decides when to weaken, create or abandon a hypothesis based on observed results.
See Appendix \ref{app:icbf} for full details.

\subsection{Baselines}\label{sub:baselines}

\paragraph{Random.} Genes selected uniformly at random without replacement (10 replicates).

\paragraph{GP-UCB.} This baseline is a Gaussian Process regression with Upper Confidence Bound acquisition, using a gene--gene similarity kernel derived from STRING protein--protein interaction scores \citep{szklarczyk2023string} (10 replicates). We chose STRING PPI scores because they provide a principled, off-the-shelf measure of gene--gene functional similarity. This is a deliberately conservative baseline: richer kernels incorporating gene expression similarity, Gene Ontology semantic similarity, or learned embeddings could yield stronger GP-UCB performance and would narrow the gap with LLM agents.

\paragraph{Random Feedback.} This baseline uses same approach as ICL-EF, but the hit/miss labels in the experimental history are randomly permuted within each batch, breaking the correlation between gene identity and outcome while preserving the marginal hit rate per iteration (10 replicates, Sonnet~4.6 only).

\subsection{Statistical significance of pairwise methods comparison across features}
We treat target features as the unit of inference. For each method and feature, performance (cumulative unique discoveries) is first averaged across the 10 independent campaign replicates, which provide repeated measurements of the same feature under identical experimental conditions. Replicates are therefore used to obtain a stable estimate of feature-level performance but are not treated as independent samples in the final significance test. Pairwise method comparisons are then performed on the resulting paired feature-level scores using an exact two-sided sign-flip permutation test across features, which tests whether the mean feature-wise performance difference could arise under the null hypothesis of exchangeable method labels. 

Because multiple pairwise method comparisons are performed, we control the false discovery rate (FDR) using the Benjamini–-Hochberg procedure applied to the set of permutation p-values across all method pairs. Unless otherwise stated, reported p-values correspond to these Benjamini–-Hochberg corrected permutation p-values. 
In addition, for further descriptive analysis, per-feature comparisons between methods using paired two-sided sign-flip permutation tests on the replicate-level performance differences within each feature are reported in Appendix \ref{app:per-feat-stats}. This appendix also shows hierarchical bootstrap confidence intervals that can account for the combined sources of variability arising from both feature-to-feature differences and stochasticity across campaign replicates.

\section{Results}

\subsection{In-Context Learning Improves Performance}

\begin{table*}[t]
\centering
\small
\caption{Cumulative discoveries across 10 target features (F0--F90). Per-feature columns report mean $\pm$ std over 10 replicates. The ``All'' column reports the mean and std of the 10 per-feature means, capturing between-feature heterogeneity. Bold indicates best per feature. 
The Sonnet~4.5 panel shows that ICL-EF fails to produce significant gains compared to zero-shot ($+0.8$, $p=0.32$), while Sonnet~4.6 yields large ICL-EF effects ($+8.9$, $p =0.003$). Moreover, ICBR-EF exhibits marginal but consistent improvement over ICL-EF ($+2.1$, $p=0.006$).}
\label{tab:full}
\begin{tabular}{@{}lccccccccccc@{}}
\toprule
Method & F0 & F10 & F20 & F30 & F40 & F50 & F60 & F70 & F80 & F90 & All \\
\midrule
Random & 18.3\tiny{$\pm$3.7} & 15.1\tiny{$\pm$4.2} & 12.8\tiny{$\pm$3.0} & 11.9\tiny{$\pm$1.9} & 10.7\tiny{$\pm$4.1} & 10.9\tiny{$\pm$2.8} & 11.8\tiny{$\pm$3.3} & 5.8\tiny{$\pm$3.0} & 8.7\tiny{$\pm$2.0} & 4.0\tiny{$\pm$1.5} & 11.0\tiny{$\pm$4.0} \\
GP-UCB & 27.1\tiny{$\pm$3.7} & 26.8\tiny{$\pm$4.3} & 23.9\tiny{$\pm$4.4} & 20.6\tiny{$\pm$4.6} & 18.9\tiny{$\pm$2.9} & 16.3\tiny{$\pm$2.9} & 17.6\tiny{$\pm$3.5} & 16.1\tiny{$\pm$3.8} & 17.3\tiny{$\pm$4.8} & 10.7\tiny{$\pm$2.2} & 19.5\tiny{$\pm$4.9} \\
\midrule
Zero-shot & 27.0\tiny{$\pm$2.4} & 20.0\tiny{$\pm$4.0} & 25.2\tiny{$\pm$3.2} & 21.1\tiny{$\pm$2.5} & 20.1\tiny{$\pm$1.8} & 17.4\tiny{$\pm$3.5} & 18.0\tiny{$\pm$2.0} & 18.8\tiny{$\pm$4.8} & 21.1\tiny{$\pm$3.7} & 12.1\tiny{$\pm$1.8} & 20.1\tiny{$\pm$3.9} \\
ICL-EF & 28.3\tiny{$\pm$5.2} & 27.6\tiny{$\pm$5.5} & 24.9\tiny{$\pm$4.2} & 18.4\tiny{$\pm$3.1} & 19.8\tiny{$\pm$6.5} & 18.4\tiny{$\pm$1.9} & 20.0\tiny{$\pm$3.3} & 16.3\tiny{$\pm$5.6} & 31.4\tiny{$\pm$6.5} & 12.9\tiny{$\pm$0.9} & 21.8\tiny{$\pm$5.6} \\
\midrule
Zero-shot & 24.7\tiny{$\pm$2.5} & 26.4\tiny{$\pm$3.3} & 27.0\tiny{$\pm$3.6} & 21.8\tiny{$\pm$2.5} & 19.2\tiny{$\pm$2.0} & 18.0\tiny{$\pm$2.1} & 17.3\tiny{$\pm$2.3} & 19.7\tiny{$\pm$1.8} & 17.0\tiny{$\pm$4.4} & 12.8\tiny{$\pm$1.3} & 20.4\tiny{$\pm$4.3} \\
Random FB & 28.6\tiny{$\pm$5.8} & 24.2\tiny{$\pm$4.2} & 19.5\tiny{$\pm$6.0} & 22.4\tiny{$\pm$3.8} & 20.0\tiny{$\pm$3.8} & 17.6\tiny{$\pm$1.7} & 17.7\tiny{$\pm$2.5} & 13.5\tiny{$\pm$3.5} & 12.5\tiny{$\pm$3.9} & 7.6\tiny{$\pm$3.1} & 18.4\tiny{$\pm$5.8} \\
ICL-EF & 32.7\tiny{$\pm$2.9} & 37.5\tiny{$\pm$9.0} & \textbf{35.7}\tiny{$\pm$2.1} & 26.8\tiny{$\pm$2.7} & 27.6\tiny{$\pm$3.3} & 21.7\tiny{$\pm$2.2} & 21.2\tiny{$\pm$1.7} & 32.7\tiny{$\pm$2.2} & 43.1\tiny{$\pm$3.6} & 14.3\tiny{$\pm$1.0} & 29.3\tiny{$\pm$8.2} \\
ICBR-EF & \textbf{33.4}\tiny{$\pm$3.8} & \textbf{48.3}\tiny{$\pm$2.6} & 35.3\tiny{$\pm$3.1} & \textbf{28.7}\tiny{$\pm$3.2} & \textbf{28.8}\tiny{$\pm$3.4} & \textbf{22.4}\tiny{$\pm$1.6} & \textbf{21.7}\tiny{$\pm$2.2} & \textbf{34.8}\tiny{$\pm$2.0} & \textbf{44.4}\tiny{$\pm$6.7} & \textbf{16.5}\tiny{$\pm$0.8} & \textbf{31.4}\tiny{$\pm$9.5} \\
\bottomrule
\end{tabular}
\end{table*}

\begin{figure*}[t]
\centering
\includegraphics[width=\textwidth]{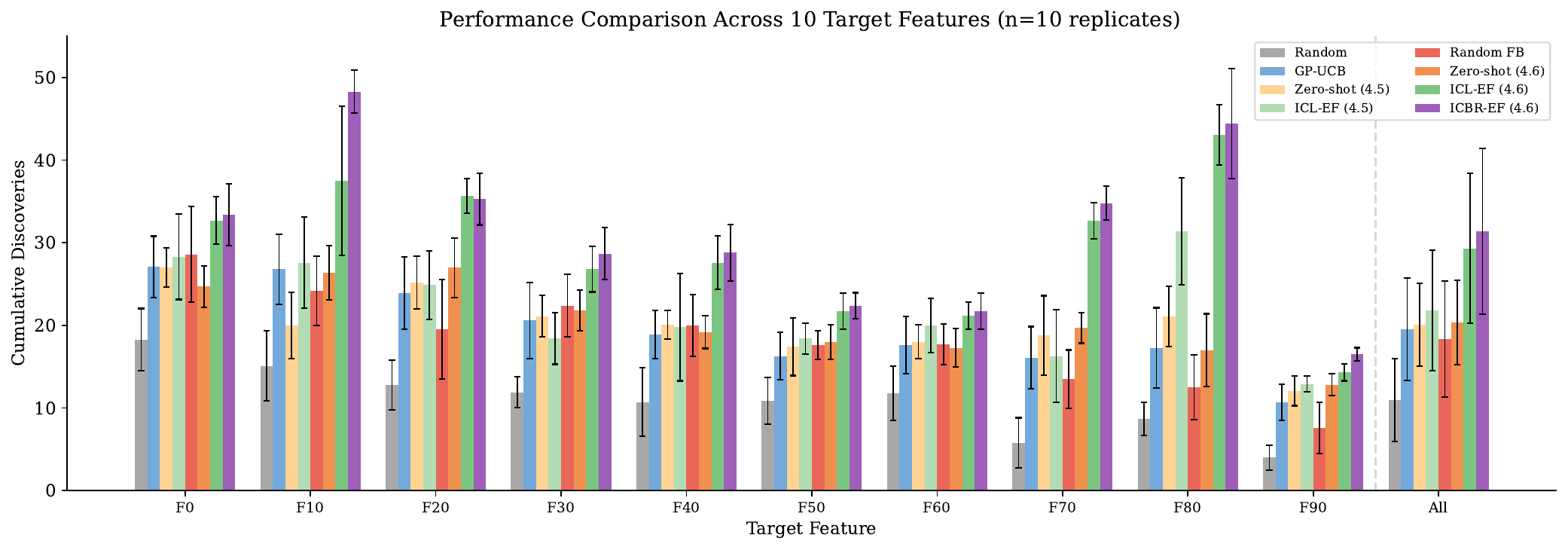}
\caption{Performance comparison of six methods across 10 target features (10 replicates per condition). Error bars show standard deviation. Random FB sits below the zero-shot agent, while ICL-EF and ICBR-EF dominate all baselines.}
\label{fig:bar}
\end{figure*}

Table~\ref{tab:full} shows a consistent improvement when Sonnet 4.6 receives explicit success/failure feedback (ICL-EF), discovering an average of $29.3$ hits compared to $20.4$ for the zero-shot agent ($p=0.003$). ICBR-EF, which receives extended phenotypic feedback, achieves the highest overall performance ($31.4$ mean discoveries) and the difference from ICL-EF is also statistically significant ($p=0.006$).

Both feedback-enabled LLMs outperform the GP-UCB BO baseline ($19.5$, $p =0.003$). This suggests that LLM-based semantic reasoning over biological literature (ingested during pretraining) can be more effective for this task than statistical regression over structural networks, though we note the GP-UCB kernel was deliberately conservative (see Subsection \ref{sub:baselines}). All methods surpass the random baseline ($11.0$ hits). 
Further method performance gap significance analyses can be found in Appendix~\ref{app:per-feat-stats}. 

\subsection{ICL-EF vs ICBR-EF: qualitative comparison}

An observation from these results is that despite the ICBR-EF agent deploying sophisticated reasoning (including phenotypic state inference and persistent hypothesis registries as analyzed in Section \ref{sec:behavior}) its ultimate quantitative gain over the simpler ICL-EF agent is modest (+2.1 discoveries).

Qualitative analysis reveals two primary reasons accounting this. First, the ICL-EF agent is highly efficient at exploitation. Once a few hits in major protein complexes (e.g., transcription machinery, nuclear envelope, or the ubiquitin-proteasome system) are found, the basic feedback mechanism allows ICL-EF to mine out virtually the entire complex before the 1,000-test budget is exhausted. 
Second, while the ICBR-EF agent's capacity to infer novel morphological connections (phenotypic clustering) successfully identifies genes across disjoint pathways, the absolute number of these pure, multi-pathway targets that are actually contained within the physical screening library is small. Thus, the additional effort of complex phenotypic inference yields only a few additional discoveries per feature, creating a potential point of diminishing returns for purely in-context learning loops.

\begin{figure}[t]
\centering
\includegraphics[width=\columnwidth]{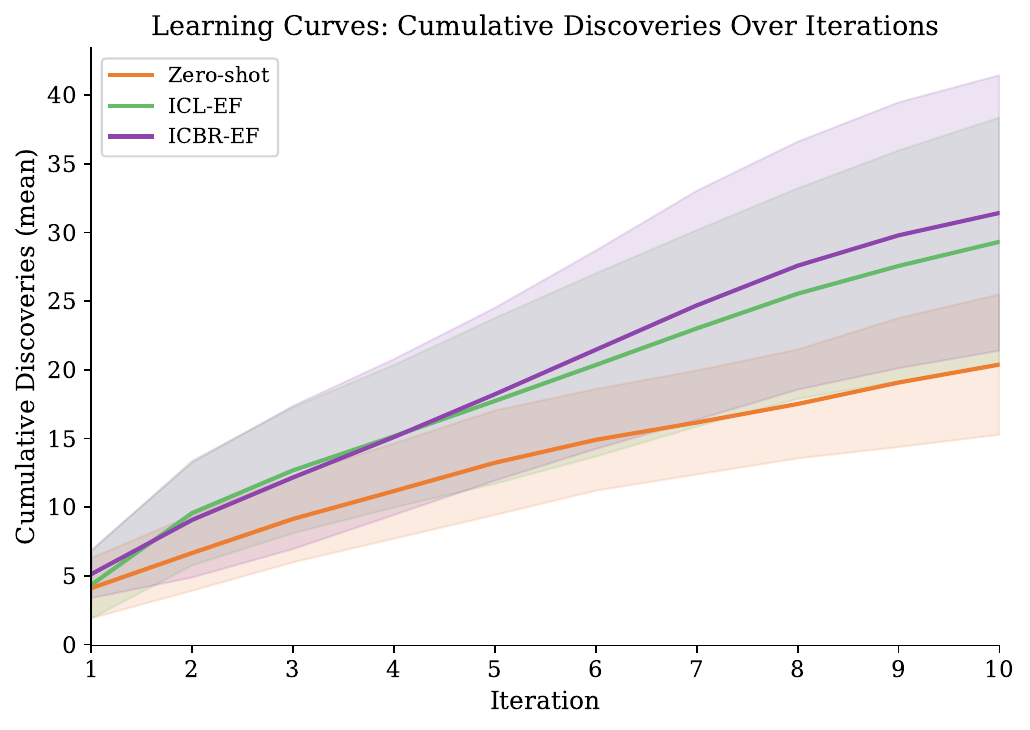}
\caption{Cumulative discoveries over 10 iterations. ICL-EF progressively outpaces the zero-shot baseline as feedback accumulates.}
\label{fig:curves}
\end{figure}

Figure~\ref{fig:curves} shows learning curves. The gap between ICL-EF and the zero-shot baseline widens progressively as observations accumulate, consistent with genuine in-context learning. By iteration 10, the ICL-EF agent has discovered $10$ more genes on average than the zero-shot baseline. 
The cumulative discoveries achieved by ICL-EF and ICBR-EF remain on par until iteration 6, after which a contrast is visible but remains marginal. 

\subsection{Feature-Level Heterogeneity}\label{sub:feature_level_heterogeneity}

\begin{figure}[t]
\centering
\includegraphics[width=\columnwidth]{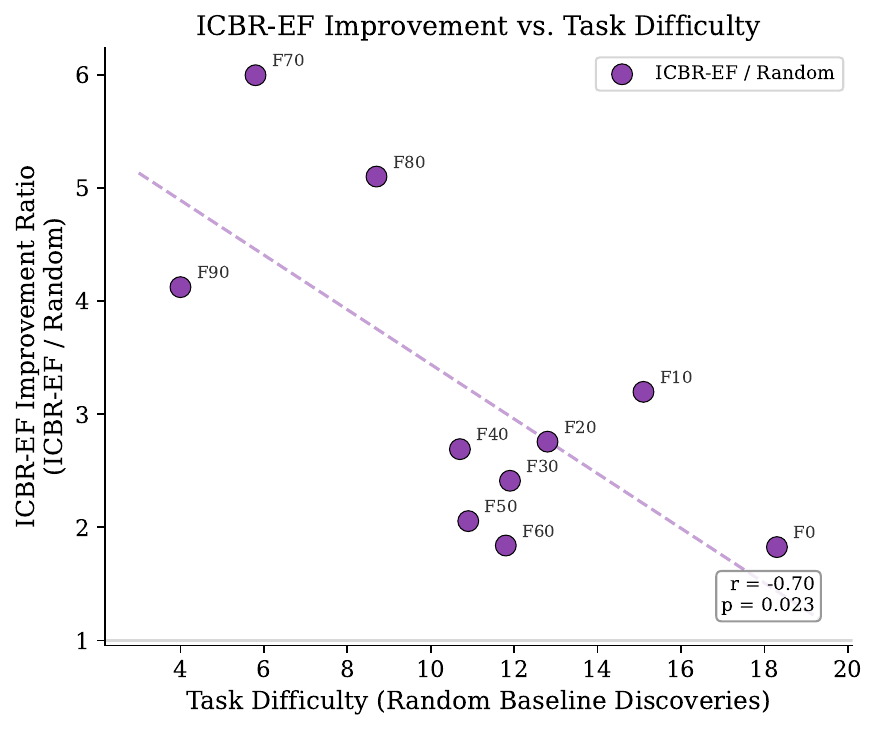}
\caption{ICBR-EF improvement vs.\ Random baseline. Values above 1.0 indicate the ICBR-EF agent consistently outperformed random selection, achieving up to a $6.0\times$ improvement on F70.}
\label{fig:scatter}
\end{figure}

Table~\ref{tab:full} and Figure~\ref{fig:scatter} reveal that while ICL is consistently beneficial, its magnitude varies substantially. Compared to the random baseline, the ICBR-EF agent's largest relative improvements appear on F70 ($6.0\times$ more discoveries), F80 ($5.1\times$), and F10 ($3.2\times$), where gene family structure allows the agent to exploit discovered hits aggressively. Even on the most difficult features like F90, learning provides a significant benefit ($4.1\times$).

This heterogeneity has practical implications. The overall effect ($+20.4$ discoveries) is largest on features where gene family structure creates learnable correlations, and small (but still significant) on features with less exploitable structure.

The higher variance of both ICL-EF (std=8.2) and ICBR-EF (std=9.5) vs for zero-shot (std=4.3) reflects this: feedback creates both productive (gene family exploitation) and unproductive (misplaced obstinacy) behavioral modes, leading to higher upside but also more variable outcomes.
 
\subsection{Model Capability as a Critical Factor}
\label{sec:model}

\begin{figure}[t] 
\centering
\begin{subfigure}{\columnwidth}
  \centering
  \includegraphics[width=\linewidth]{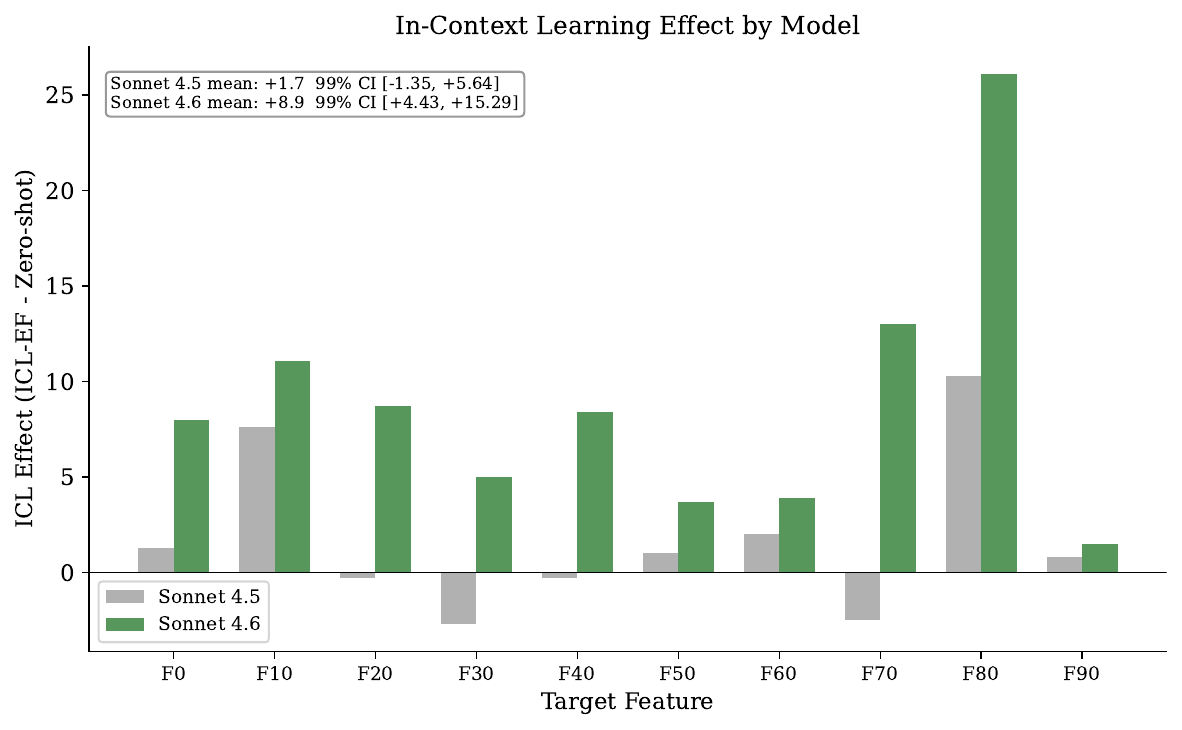}
  \caption{Per-feature ICL effect (ICL-EF $-$ Zero-shot) for Sonnet~4.5 vs.\ 4.6. With 4.5, ICL is inconsistent (5/10 positive); with 4.6, it is universally positive.}
\end{subfigure}
\vspace{4pt}
\begin{subfigure}{\columnwidth}
  \centering
  \includegraphics[width=\linewidth]{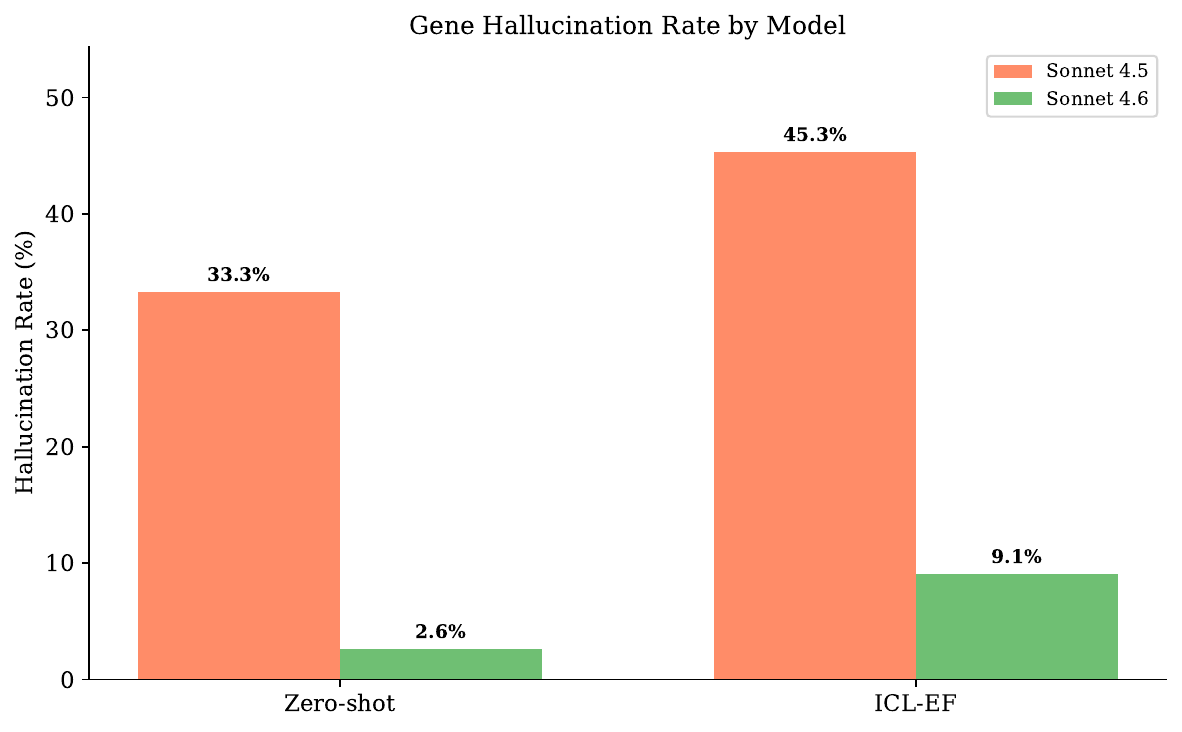}
  \caption{Gene hallucination rate drops from ${\sim}33\%$--$45\%$ to ${\sim}3$--$9\%$.}
\end{subfigure}
\caption{Effect of model capability on ICL and hallucination.}
\label{fig:model}
\end{figure}

Table~\ref{tab:full} and Figure~\ref{fig:model} present a direct comparison between Claude Sonnet~4.5 and~4.6 using the same agent architecture and prompts.
Based on per-feature p-values and effect confidence intervals (CIs) presented in Appendix~\ref{app:per-feat-stats}, the contrast is stark: with Sonnet~4.5, the ICL effect is only $+1.7$ discoveries (99\% effect CI $ = [-2.03, +6.40]$, not significant), positive on 5/10 features. With Sonnet~4.6, the ICL effect is $+8.9$ (99\% effect CI $ = [+4.50, +15.33]$, significant) for ICL-EF, positive on 10/10 features, and significant at $p < 0.01$ on 7 of them. 
As expected, the effect is stronger with ICBR-EF (99\% effect CI $ = [+5.81, +17.93]$, significant), positive and significant positive on 10/10 features.

One measurable mechanism is \textbf{gene hallucination}. By parsing the LLM's raw completions against the valid gene library (${\sim}8{,}000$ JUMP genes), we measure the fraction of proposed genes that do not exist in the screening library:

\begin{itemize}
\item \textbf{Sonnet 4.5}: 45.3\% hallucination rate, 39--50\% of test slots filled by random fallback (interquartile range across replicates);
\item \textbf{Sonnet 4.6}: 9.1\% hallucination rate, 3--11\% random fallback (interquartile range across replicates).
\end{itemize}

\noindent Importantly, the ``hallucinated'' genes are not random strings, i.e. they are real biological genes (e.g., EXOSC1, KDM3A, MED18) from the correct gene families, but absent from the ${\sim}8{,}000$-gene JUMP screening library. The failure is one of \emph{instruction-following}: the model ignores the provided list of available (untested) genes and instead proposes genes from its own biological knowledge. This worsens over iterations (from $0\%$ at iteration~1 to ${\sim}60\%$ at iteration~6 plateauing at this level in the remaining iterations) as the untested list shrinks and the model increasingly defaults to its priors. Sonnet~4.6's improvement reflects better instruction-following. It reliably constrains its output to the provided gene list. The model upgrade produces a ${\sim}5\times$ reduction in out-of-library proposals and a corresponding $34\%$ increase in absolute discoveries (from 21.8 to 29.3 for ICL-EF).

However, a model upgrade changes multiple capabilities simultaneously (world knowledge, reasoning quality, instruction adherence, hallucination rate, etc.) so the improvement cannot be attributed to hallucination reduction alone. The Sonnet~4.6 model may also have better biological priors or better ability to process long contexts. What we can conclude is that sufficient model capability is necessary for ICL to manifest: studies using earlier or weaker models may have failed to observe ICL partly because hallucination prevents the model from executing its reasoning.

\subsection{Behavioral Analysis: The Evolution of Search Strategies}
\label{sec:behavior}

Analysis of the agents' reasoning traces reveals exactly how experimental feedback and phenotypic context drive the stepwise jumps in performance. An in-depth analysis is proposed in Appendices~\ref{app:simple} to \ref{app:icbf}. 
Overall, we identify a three-tiered evolution in search strategy:

\paragraph{Zero-shot vs.\ ICL-EF: From Static Priors to Pathway Exploitation.}
The zero-shot agent relies entirely on its biological priors, making static best-guesses (e.g., broadly nominating ``actin regulators'' or ``kinases''). It cannot adapt. In contrast, the ICL-EF agent uses binary hit/miss feedback to confirm an initial guess, and then aggressively pivots. If it discovers a single hit in the SWI/SNF complex, it immediately targets the remaining untested members of that specific complex. Semantic analysis of completion logs shows ICL-EF uses terminology related to targeted ``complex/family/subunit'' exploitation an average of 7.9 times per iteration, versus 2.3 for the zero-shot agent (a $3.4\times$ higher rate consistent with active exploitation of experimental feedback). This allows it to systematically mine horizontal literature connections that zero-shot approaches cannot see.

\paragraph{ICL-EF vs.\ ICBR-EF: From Pathway Exploitation to Phenotypic State Inference.}
While powerful, ICL-EF is fundamentally limited to proposing genes that share a known literature or structural relationship to a confirmed hit. 
The ICBR-EF agent transcends this limitation. Because its prompt includes a rich, multidimensional phenotypic fingerprint (the top 10 most significant feature perturbations for each tested gene), it performs \textit{phenotypic state inference}. Rather than just grouping genes by name, it clusters hits by their multi-dimensional cellular effects. 

For example, on F80, the ICBR-EF agent recognized that hits spanning disjoint functional pathways (e.g., DNA damage repair, cyclin-CDK complexes, and mitotic kinases) all produced identical perturbations in ``angular second moment'' and ``nuclear texture.'' The agent correctly deduced these diverse targets converged on the same underlying physiological state: mitotic arrest. Semantic analysis confirms this strategic shift: the ICBR-EF agent explicitly references phenotypic markers and cellular states 22.6 times per iteration (compared to just 0.7 times for binary ICL-EF). By clustering genes by their \textit{effects} rather than their \textit{names}, the ICBR-EF agent can jump across disconnected biological pathways to discover novel, non-obvious hits.

\paragraph{Stateful Tracking vs.\ Reactive Prompting.}
Finally, tracing the evolution of the ICBR-EF agent's \textit{hypotheses register} reveals a critical architectural advantage. While both agents receive the cumulative history of all tested genes and their outcomes, only the ICBR-EF agent maintains an explicit theory layer on top of this history. The ICL-EF agent re-derives its strategy from scratch at each iteration: it sees the full observation log but has no mechanism to preserve the \emph{reasoning} behind previous selections. Over 10 iterations, the interpretation of early hits can drift as newer results dominate the prompt, and there is no way to mark a line of inquiry as exhausted. The ICBR-EF agent, by contrast, carries forward a structured hypotheses register across iterations, explicitly tracking the lifecycle of its theories (tagging them as \textit{Active}, \textit{Confirmed}, or \textit{Weakened}) alongside explicit justifications (e.g., ``\textit{PSMB8,9,10 (immunoproteasome) are MISSes, confirming specificity to constitutive proteasome}''). This persistent reasoning scaffold prevents interpretive drift and allows the agent to systematically close out search spaces rather than revisiting them. Ultimately, this combination of phenotypic state inference and persistent hypothesis tracking drives the ICBR-EF agent's superior performance.

\subsection{Reconciling with Prior Work}\label{sub:reconciling_with_prior_work}

\begin{figure}[t]
\centering 
\includegraphics[width=\columnwidth]{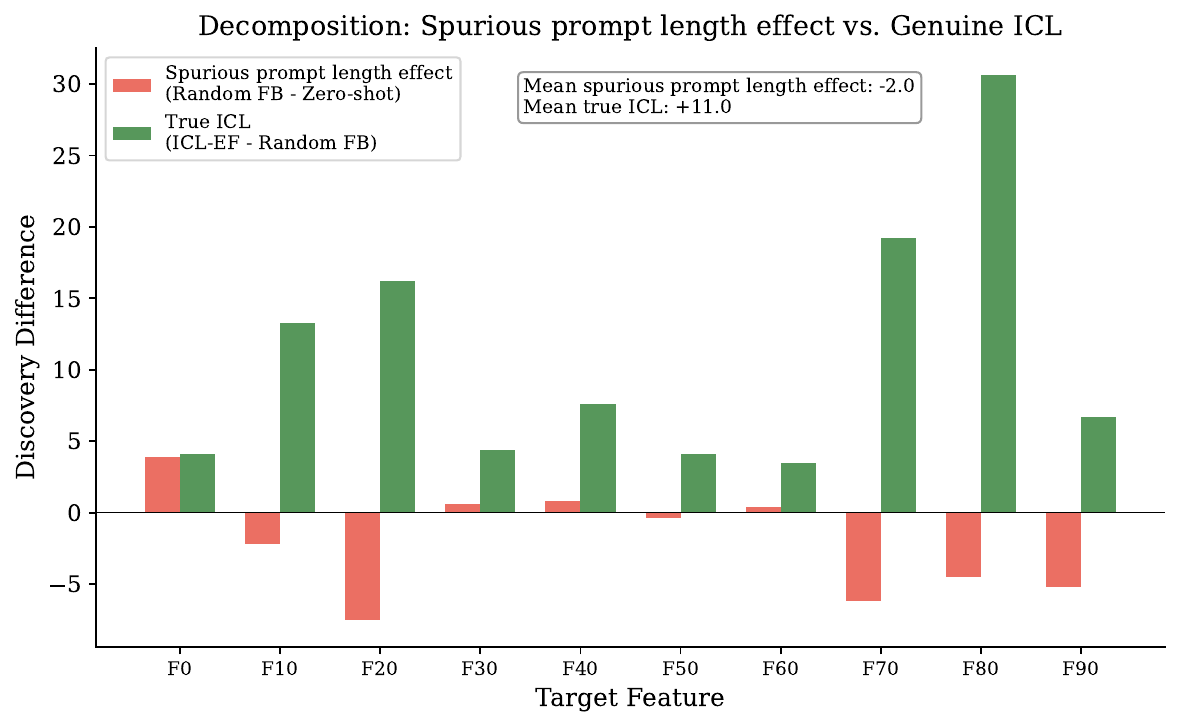}
\caption{Decomposition of the learning effect into memory jogging (Random FB $-$ Zero-shot, red) and genuine ICL (ICL-EF $-$ Random FB, green) per feature with Sonnet 4.6. The spurious prompt length effect is negative on 6/10 features; genuine ICL is positive on all 10.}
\label{fig:decomp}
\end{figure}

Why do we observe significant ICL where \citet{gupta2025llms} report none? To directly test their hypothesis, i.e. that agents spuriously benefit from the prompt's length rather than its factual content, we ran a \textbf{Random Feedback} (Random FB) control across all 10 target features (10 replicates each, 100 experiments total). In this condition, the agent receives an identically structured prompt containing the full experimental history, but the hit/miss labels are randomly permuted among the tested genes within each batch. This within-batch permutation preserves the marginal hit rate per iteration while breaking the correlation between gene identity and outcome. An alternative scheme (permuting labels across iterations) would preserve per-gene labels but scramble temporal ordering, testing for a different confound. Our scheme directly tests whether the agent uses the \emph{content} of feedback (which gene was a hit) rather than its \emph{structure} (how many hits occurred).

The results do not support the (pretraining) knowledge retrieval hypothesis. Averaged across all 10 features, the zero-shot agent discovered $20.4$ hits. The Random FB agent, receiving permuted outcomes, discovered only $18.4$ hits, on par or worse than no feedback at all ($-2.0$, $p=0.15$ not significant). A true feedback agent such as ICL-EF discovered 29.3 hits, a $+10.9$ improvement over Random FB ($p = 0.003$). 
ICBR-EF discovered 31.4 hits, a $+13.0$ improvement over Random FB ($p = 0.003$).

Figure~\ref{fig:decomp} reveals that the randomization effect is negative on 6/10 features and significantly harmful on 4 (F20, F70, F80, F90).
This reveals a notable asymmetry: random feedback actively harms the agent's reasoning. When the agent receives false positive signals, it pursues unproductive gene families (a form of misplaced obstinacy effect), wasting its budget on misleading leads. On the hardest features (F70, F80, F90), where exploitable structure is sparse, random feedback reduces discoveries by $30{-}40\%$ relative to no feedback. The agent is not merely reading the prompt but and is acting on the specific factual content, and false content produces systematically worse outcomes than no content at all.

We identify several factors explaining the divergence from \citet{gupta2025llms}:

\begin{enumerate}
\item \textbf{Model capability.} As shown in Section~\ref{sec:model}, ICL is not significant with Sonnet~4.5 ($p=0.55$) but highly significant with Sonnet~4.6 ($p =0.003$). The $5\times$ reduction in hallucination enables the agent to execute its reasoning. \citeauthor{gupta2025llms} primarily evaluated open-source and earlier Claude models, which may have lacked the requisite fidelity.
\item \textbf{Dataset structure.} JUMP Cell Painting data contains rich gene family correlations that create learnable structure for ICL exploitation.
\item \textbf{Statistical power.} With 10 replicates per condition (800 total experiments), we have sufficient power to detect the effect. 
\end{enumerate}

\section{Future Work}

ICL has inherent limitations: the agent must fit all observations into a fixed context window, learning is ephemeral (reset each session), and the agent is never explicitly trained on experimental outcomes. Reinforcement learning offers a natural extension. The JUMP dataset provides a simulated environment for RL training (Reinforcement Learning from Verifiable Rewards), and the gene family correlation or misplaced obstinacy patterns we observe map directly onto the exploration--exploitation trade-off that RL is designed to optimize. An RL-trained agent could also learn transferable strategies across feature types. We leave this direction for future investigation.

\section{Discussion}

\paragraph{Practical implications.}
Our results suggest that LLM agents can provide benefits for iterative experimental design when model capability is sufficient. The magnitude of improvement varies across features but the effect is consistently positive in our setting. Practitioners should expect the largest gains where gene family structure creates learnable correlations. Model capability matters: upgrading from Sonnet 4.5 to 4.6 reduced hallucination from ${\sim}33\%$--$45\%$ to ${\sim}3$--$9\%$, corresponding to a 34\% increase in discoveries for the ICL-EF agent.

\paragraph{Cost considerations.}
Each 10-iteration campaign requires 10 LLM API calls (one per iteration), consuming on average ${\sim}410{,}000$ input tokens and ${\sim}19{,}000$ output tokens. At current Claude Sonnet pricing (\$3/MTok input, \$15/MTok output), this yields a cost of ${\sim}$\$1.50 per campaign (\$1.41 for zero-shot, \$1.46 for ICL-EF, \$1.77 for the ICBR-EF agent). The ICBR-EF agent's higher cost reflects its longer outputs (33K vs.\ 14K tokens per campaign). For the 600 LLM-based campaigns in this study, total API cost was approximately \$900. At ${\sim}$30 discoveries per campaign (ICL-EF), the cost per discovery is roughly \$0.05. GP-UCB requires no API calls but does require precomputing the gene--gene kernel matrix. Batch API pricing (50\% discount) would reduce LLM costs to ${\sim}$\$450 total.

\paragraph{Data contamination.}
Frontier LLMs are trained on large web corpora that may include papers discussing the JUMP dataset, Cell Painting protocols, or specific gene--phenotype associations measured in JUMP. This means the LLM's ``prior knowledge'' may partially derive from having seen analyses of the same dataset during pretraining, rather than from general biological knowledge alone. While our Random Feedback control confirms that the ICL effect depends on the \emph{content} of feedback (not just static priors), the absolute performance of the zero-shot agent (prior-only) may be inflated by data contamination. This concern applies broadly to any LLM-based scientific agent evaluated on public datasets.

\paragraph{Relationship to LLMNN.}
Our work argues against the interpretation of \citet{gupta2025llms} that LLMs cannot learn from feedback but does not directly evaluate their proposed LLMNN method, which delegates iterative updates to a classical nearest-neighbor algorithm while using the LLM only for prior-based initialization. A direct comparison between our ICL approach and LLMNN on the same benchmark would be informative: LLMNN may offer advantages in settings where model hallucination is high or feedback is noisy, while pure ICL may be preferred when model capability is sufficient. We leave this comparison for future work.

\paragraph{Summarized limitations.}
(1)~We evaluate only two Claude models (Sonnet 4.5 and 4.6); results may not generalize to open-source or other proprietary models, and the improvement between versions highlights strong model-dependence. The Sonnet 4.5 $\to$ 4.6 comparison is confounded: the upgrade changes world knowledge, reasoning quality, instruction adherence, and hallucination simultaneously, so the ICL improvement cannot be attributed to any single factor.
(2)~All experiments use pre-computed p-values from the JUMP dataset. Real experimental campaigns involve batch effects, measurement noise, and variable hit rates across plates that could substantially alter ICL dynamics. The clean, binary, noise-free feedback in our setup may be particularly favorable for ICL; real experiments would introduce stochasticity that could erode the advantage. How robust ICL is to noisy or delayed feedback remains an open question.
(3)~Cell Painting with CellProfiler features represents one assay type with specific correlation structures.
(4)~Our Random FB control uses within-batch permutation; across-iteration permutation would test a different confound (see Section \ref{sub:reconciling_with_prior_work}).
(5)~We do not evaluate the LLMNN hybrid method proposed by \citet{gupta2025llms}, limiting our ability to compare pure ICL against their proposed solution.

\section{Conclusion}

Across 800 independently replicated experiments, we find that a frontier LLM agent can learn from experimental feedback for iterative perturbation discovery. The feedback-enabled agent achieves $31.4$ mean discoveries ($+185\%$ over random), with the ICL effect significant at $p =0.003$. A random feedback control shows the learning effect is attributable to ICL ($+13.0$ for our best agent over permuted feedback, $p = 0.003$), with random feedback actually harming performance. Model capability is a prerequisite: the ICL effect is non-significant with Sonnet~4.5 (45.3\% hallucination) but highly significant with Sonnet~4.6 (9.1\% hallucination). These results are specific to one dataset (JUMP Cell Painting), one assay type, and two proprietary models using retrospective data. Whether these findings generalize to other experimental domains, noisier feedback settings, and open-source models remains to be established.

{\small
\bibliography{references}

@article{szklarczyk2023string,
  title={The STRING database in 2023: protein--protein association networks and functional enrichment analyses for any sequenced genome of interest},
  author={Szklarczyk, Damian and Kirsch, Rebecca and Koutrouli, Mikaela and Nastou, Katerina and Mehryary, Farrokh and Hachilif, Radja and Gable, Annika L and Fang, Tao and Doncheva, Nadezhda T and Pyysalo, Sampo and others},
  journal={Nucleic acids research},
  volume={51},
  number={D1},
  pages={D483--D489},
  year={2023},
  publisher={Oxford University Press}
}

@misc{anthropic2025claude,
  title={Claude Sonnet 4 Model Family},
  author={{Anthropic}},
  year={2025},
  howpublished={\url{https://www.anthropic.com/claude}},
  note={Model identifiers: \texttt{claude-sonnet-4-5-20250929} and \texttt{claude-sonnet-4-6}. Accessed: 2025-11-25}
}

@article{lu2025cellclip,
  title={CellCLIP--Learning Perturbation Effects in Cell Painting via Text-Guided Contrastive Learning},
  author={Lu, Mingyu and Weinberger, Ethan and Kim, Chanwoo and Lee, Su-In},
  journal={arXiv preprint arXiv:2506.06290},
  year={2025}
}

@article{wang2024can,
  title={Can In-context Learning Really Generalize to Out-of-distribution Tasks?},
  author={Wang, Qixun and Wang, Yifei and Wang, Yisen and Ying, Xianghua},
  journal={arXiv preprint arXiv:2410.09695},
  year={2024}
}

@article{falck2024context,
  title={Is in-context learning in large language models bayesian? a martingale perspective},
  author={Falck, Fabian and Wang, Ziyu and Holmes, Chris},
  journal={arXiv preprint arXiv:2406.00793},
  year={2024}
}

@article{chen2023evoprompting,
  title={Evoprompting: Language models for code-level neural architecture search},
  author={Chen, Angelica and Dohan, David and So, David},
  journal={Advances in neural information processing systems},
  volume={36},
  pages={7787--7817},
  year={2023}
}

@article{ramos2023bayesian,
  title={Bayesian optimization of catalysts with in-context learning},
  author={Ramos, Mayk Caldas and Michtavy, Shane S and Porosoff, Marc D and White, Andrew D},
  journal={arXiv preprint arXiv:2304.05341},
  year={2023}
}

@article{liu2024large,
  title={Large language models to enhance bayesian optimization},
  author={Liu, Tennison and Astorga, Nicol{\'a}s and Seedat, Nabeel and van der Schaar, Mihaela},
  journal={arXiv preprint arXiv:2402.03921},
  year={2024}
}

@article{zhang2025large,
  title={Large language models to accelerate organic chemistry synthesis},
  author={Zhang, Yu and Han, Yang and Chen, Shuai and Yu, Ruijie and Zhao, Xin and Liu, Xianbin and Zeng, Kaipeng and Yu, Mengdi and Tian, Jidong and Zhu, Feng and others},
  journal={Nature Machine Intelligence},
  pages={1--13},
  year={2025},
  publisher={Nature Publishing Group UK London}
}

@article{wang2025polo,
  title={POLO: Preference-Guided Multi-Turn Reinforcement Learning for Lead Optimization},
  author={Wang, Ziqing and Wen, Yibo and Pattie, William and Luo, Xiao and Wu, Weimin and Hu, Jerry Yao-Chieh and Pandey, Abhishek and Liu, Han and Ding, Kaize},
  journal={arXiv preprint arXiv:2509.21737},
  year={2025}
}

@article{abhyankar2025accelerating,
  title={Accelerating Materials Design via LLM-Guided Evolutionary Search},
  author={Abhyankar, Nikhil and Kabra, Sanchit and Desai, Saaketh and Reddy, Chandan K},
  journal={arXiv preprint arXiv:2510.22503},
  year={2025}
}

@article{boiko2023autonomous,
  title={Autonomous chemical research with large language models},
  author={Boiko, Daniil A. and MacKnight, Robert and Kline, Ben and Gomes, Gabe},
  journal={Nature},
  volume={624},
  number={7992},
  pages={570--578},
  year={2023},
  publisher={Nature Publishing Group}
}

@article{bray2016cell,
  title={Cell Painting, a high-content image-based assay for morphological profiling using multiplexed fluorescent dyes},
  author={Bray, Mark-Anthony and Singh, Shantanu and Han, Han and Davis, Chad T. and Borgeson, Blake and Hartland, Chris and others},
  journal={Nature Protocols},
  volume={11},
  number={9},
  pages={1757--1774},
  year={2016},
  publisher={Nature Publishing Group}
}

@article{chandrasekaran2023jump,
  title={JUMP Cell Painting dataset: morphological impact of 136,000 chemical and genetic perturbations},
  author={Chandrasekaran, Sinu N. and Ceulemans, Hugo and Boyd, Joseph D. and Carpenter, Anne E. and Kalinin, Alexandr},
  journal={bioRxiv},
  year={2023}
}

@inproceedings{gupta2025llms,
  title={LLMs for Bayesian Optimization in Scientific Domains: Are We There Yet?},
  author={Gupta, Rishi and Hartford, Jason and Liu, Bryan},
  booktitle={Findings of EMNLP 2025},
  year={2025}
}

@inproceedings{roohani2024biodiscoveryagent,
  title={BioDiscoveryAgent: An AI Agent for Designing Genetic Perturbation Experiments},
  author={Roohani, Yusuf and Lee, Andrew and Huang, Qing and Vora, Jory and Steinhart, Zach and Huang, Kevin and Marson, Alexander and Liang, Percy and Leskovec, Jure},
  booktitle={ICLR 2025},
  year={2024}
}
}

\appendix
\onecolumn

\section{Zero-shot Agent: Prompting Strategy and Execution Trace}
\label{app:simple}

The zero-shot agent is our zero-shot LLM baseline. Unlike ICL-EF and ICBF-EF, it receives \emph{no experimental feedback}: previous test results are recorded internally for evaluation purposes but are never included in the prompt. Every iteration, the LLM selects the next batch of 100 genes using only its biological prior knowledge and the list of untested gene names.

\subsection{System Prompt}

The system prompt is fixed across all iterations:

\begin{lstlisting}
You are an expert computational biologist AI agent discovering gene perturbations
that induce significant cellular responses.

TASK: Find gene knockouts where p-value for 'Nuclei_Intensity_MeanIntensity_AGP' is < 0.05.

## Strategy - Use Biological Knowledge

You must select genes based ONLY on your biological knowledge:

1. Gene families: Genes with similar names often share functions
   - Transporter families
   - Complex subunit families
   - Modifier enzyme families

2. Functional categories relevant to nuclear phenotypes:
   - Chromatin regulators
   - Nuclear envelope proteins
   - Transcription machinery
   - DNA repair genes

3. Known phenotype associations:
   - Think about which gene categories typically affect Nuclei features

## Available Data:
- Total perturbations: 7975
- Target: Nuclei_Intensity_MeanIntensity_AGP

IMPORTANT: You have NO information about previous test results.
Select genes based purely on biological reasoning.
\end{lstlisting}

\subsection{Per-Iteration User Prompt}

At each iteration $t$, the user prompt contains three components and nothing else.

\paragraph{1. Iteration counter.} A single line \texttt{Iteration $t$} provides context on how far into the experiment the agent is.

\paragraph{2. Already-tested and untested gene lists.} The full list of already-tested gene IDs (to prevent repeats) and the full list of untested IDs are provided verbatim.

\paragraph{3. Self-check and output format.} Identical to ICL-EF and ICBF-EF:

\begin{lstlisting}
## Iteration 2

Select 100 genes to test from the untested perturbations below.

Use your biological knowledge to prioritize genes most likely to affect
'Nuclei_Intensity_MeanIntensity_AGP'.

ALREADY TESTED PERTURBATIONS (DO NOT SUGGEST THESE):
ARID1A, ATM, ATR, ...   [100 gene names]

UNTESTED PERTURBATIONS (7875 remaining):
A2M, A3GALT2, A4GALT,  ...   [7875 gene names]

CRITICAL REQUIREMENT:
Before returning your final JSON, you MUST explicitly double-check your proposed
genes ONE BY ONE. For each gene you want to propose, verify:
1. It is EXACTLY present in the UNTESTED list above.
2. It is NOT present in the ALREADY TESTED list.
If you realize a gene violates these rules, propose a replacement.

After your verification, return the final 100 genes strictly inside a JSON code block:
```json
["GENE1", "GENE2", "GENE3", ...]
```
\end{lstlisting}


\subsection{Execution Trace}
We trace the agent on \texttt{Nuclei\_Intensity\_MeanIntensity\_AGP} / F0 (152 significant genes out of 7,975), with batch size 100 and 10 iterations (1,000 total tests).

\paragraph{Iteration 1 --- Biology-prior initialization.}
With no feedback available, the LLM first interprets the target feature: the AGP channel (actin, Golgi, plasma membrane) measures cytoskeletal staining that can overlap with nuclei, so nuclear intensity in this channel is expected to be sensitive to nuclear envelope/lamina integrity, chromatin compaction, cell cycle state, histone modifications, and transcription factor localisation. It then selects 100 genes organised by these mechanisms: chromatin regulations and remodeling (\texttt{HDAC1--3}, \texttt{EZH2}, \texttt{KDM1A}, \texttt{KDM5B}, \texttt{SETD2}, \texttt{CHD4}), nuclear transport and nuclear envelope integrity maintenance (\texttt{LMNA}, \texttt{LBR}, \texttt{NUP98}, \texttt{NUP214}), cell cycle (\texttt{CDK1/2/4}, \texttt{CCNA2}, \texttt{PLK1}, \texttt{AURKA/B}), DNA-damage checkpoints (\texttt{ATM}, \texttt{ATR}, \texttt{BRCA1/2}, \texttt{PCNA}), and transcriptional regulation  (\texttt{MYC}, \texttt{EP300}, \texttt{TP53}, \texttt{MDM2}). It also explicitly verifies each selected gene against the untested pool before outputting the final list.
Hits: \texttt{CHD4, PCNA, MYC, EP300, MDM2} (5 of 100).

\paragraph{Iteration 2 --- Continued prior-based selection.}
Still operating without experimental feedback, the agent broadens its prior-based sweep across various mechanistic categories: remaining KDM/HAT/HDAC family members (\texttt{KDM1B}, \texttt{KDM5A/C/D}, \texttt{KDM7A}, \texttt{KAT6A/B}, \texttt{KAT7/8}, \texttt{HDAC4--9}), SET-domain writers (\texttt{EZH1}, \texttt{EHMT1}, \texttt{SETD7}, \texttt{SETDB2}), nuclear transport (\texttt{KPNA1/2}, \texttt{KPNB1}, \texttt{XPO1}), extended CDK family (\texttt{CDK5--9}, \texttt{CDK12}), transcription factors (\texttt{FOXM1}, \texttt{FOXO1/3}, \texttt{RUNX1/2}, \texttt{STAT3}), chromatin readers and remodellers (\texttt{BRD2--4}, \texttt{PHF} proteins, \texttt{UHRF1/2}, \texttt{TRIM28/33}), CHD-family remodellers (\texttt{CHD1/2/7/8}), and signalling (\texttt{MTOR}).
Hits: \texttt{XPO1, TRRAP, UHRF1, KMT2B} (4 of 100).

\paragraph{Iterations 3--10 --- Exhaustive prior coverage without adaptation.}
Across the remaining eight iterations the agent's strategy does not evolve: lacking any feedback, it restarts the same biological reasoning from scratch each round, re-identifying the same four to five categories (transcription factors, chromatin remodellers, nuclear transport, cytoskeletal and kinase regulators) and simply advancing through the untested pool within those categories.
Hit rates remain low but non-zero (iter.\ 3: 4; 4: 1; 5: 3; 6: 1; 7: 6; 8: 2; 9: 1; 10: 1), reflecting a gradually depleted prior rather than learned focus.
The spike in iteration~7 arises incidentally when the agent reaches RNA-polymerase and Mediator subunits (\texttt{POLR2A}, \texttt{MED21}, \texttt{MED27}) by exhausting adjacent chromatin-remodelling families, not by inference from earlier hits.
Because results are never fed back, the agent cannot learn that its hits are enriched in the transcription-coactivation cluster and therefore fails to efficiently target related machinery in later iterations.

\paragraph{Summary.}
After 1,000 tests the agent discovers 28 of 152 significant genes (18.4\% recall): \texttt{CHD4, PCNA, MYC, EP300, MDM2, XPO1, TRRAP, UHRF1, TRIM28, KMT2B, SUPT6H, TAF1, TAF5, AURKC, CBFA2T3, PSMC2, IRF3, RAC1, POLR2A, MED21, MED27, RBM14, PRPF19, CAND1, ZBTB45, ZSCAN9, ATP6V0D1, KIF11}. 
Hit rate drops after iteration~3 as the agent exhausts its highest-confidence prior categories, then partially recovers in iteration~7 when it independently reaches new gene families. Because results are never fed back, the agent cannot exploit for example exploit the CHD4/EP300 hits from iteration~1 to suggest TAF1 in iteration~2 which is also an epigenetic regulator gene.

\section{ICL-EF Agent: Prompting Strategy and Execution Trace}
\label{app:icl}

The ICL-EF (In-Context Learning with Experimental Feedback) agent is the core feedback-enabled agent in our study. At each iteration the LLM receives the full outcome history of all tested genes (hit/miss and target p-value) and uses its biological prior knowledge to select the next batch. 
To guide the exploitation of successful hits, we also provide gene name patterns. We extract the leading alphabetic characters (up to four) of each successful gene name, count how often each prefix recurs across all hits, and expose the five most frequent prefix–count pairs to the LLM so that it can immediately identify which gene families are over-represented among successes and prioritize untested members of those families in the next batch.

\subsection{System Prompt}

The system prompt primes the model for gene-family and pathway reasoning and is fixed across all iterations:

\begin{lstlisting}
You are an expert computational biologist AI agent discovering gene perturbations
that induce significant cellular responses.

TASK: Find gene knockouts where p-value for
      'Nuclei_Intensity_MeanIntensity_AGP' is < 0.05.

KEY STRATEGY - Use your biological knowledge:
1. Gene families: Genes with similar names often have related functions
2. Pathways: If a gene in a pathway shows effect, test other genes in same pathway
3. Protein complexes: Subunits of same complex often show similar phenotypes
4. Functional categories: Transcription factors, kinases, transporters may cluster

PATTERN RECOGNITION:
- When successful genes share naming patterns or prefixes, prioritize similar names
- Look for gene family numbers - if one member works, test other family members
- Consider biological process: if mitochondrial genes work, test more mitochondrial
  genes

AVAILABLE DATA:
- Total perturbations: 7975
- Features: 100 CellProfiler measurements (p-values)
- Target: Nuclei_Intensity_MeanIntensity_AGP

Remember: You're looking for genes affecting nuclear morphology/intensity.
Think about:
- Chromatin regulators (histones, KMTs, KDMs, CHDs)
- Nuclear envelope proteins
- Transcription machinery (TAFs, POLRs)
- Signaling to nucleus
\end{lstlisting}

\subsection{Per-Iteration User Prompt}

The user prompt is assembled from four components at each iteration~$t$.

\paragraph{1. Observation summary.} All tested genes are listed with a hit/miss label and their target p-value:

\begin{lstlisting}
TESTED PERTURBATIONS: 100
Successes: 8, Failures: 92

RESULTS:
  CHD1 :  MISS (p=0.6244)
  CHD2 :  MISS (p=0.6439)
  ...
  CHD4 :  HIT  (p=0.0004)
  KMT2  : HIT  (p=0.0024)
  TRRAP:  HIT  (p=0.0308)
  MYC. :  HIT  (p=0.0070)
  ...
\end{lstlisting}

\paragraph{2. Gene prefix summary.} The framework extracts the four leading alphabetic characters of each hit gene name, counts prefix frequencies, and appends the top-5 as a compact hint:

\begin{lstlisting}
SUCCESSFUL GENE PATTERNS: CHD(1), KMTB(1), EP(1), TAF(1), POLR(1)
\end{lstlisting}

\paragraph{3. Already-tested and untested gene lists.} Full ID lists so the agent can verify its proposals do not re-test any gene.

\paragraph{4. Strategy and self-check.} Instructs the agent to exploit observed hits, explore novel categories, and verify each proposed gene before returning the final list:

\begin{lstlisting}
STRATEGY FOR SELECTION:
1. EXPLOIT: Find genes with similar names/prefixes to successful ones
2. EXPLOIT: Test genes in same pathway/complex as successes
3. EXPLORE: Include some genes from untested categories for diversity

CRITICAL REQUIREMENT:
Before returning your final JSON, you MUST explicitly double-check your proposed
genes ONE BY ONE. For each gene you want to propose, verify:
1. It is EXACTLY present in the UNTESTED list above.
2. It is NOT present in the ALREADY TESTED list.
If you realize a gene violates these rules, propose a replacement.

After your verification, return the final 100 genes strictly inside a JSON
code block:
```json
["GENE1", "GENE2", "GENE3", ...]
```
\end{lstlisting}

\subsection{Execution Trace}

We trace the agent on \texttt{Nuclei\_Intensity\_MeanIntensity\_AGP} / F0 (152 significant genes out of 7,975), batch size 100, 10 iterations (1,000 total tests).

\paragraph{Iteration 1 --- Prior-guided initialisation.}
With no observations yet, the agent draws on its biological prior and explicitly lists nine mechanistic categories: chromatin remodelers (\texttt{CHD1/2/4/7/8}, full SWI/SNF complex); histone methyltransferases and demethylases (\texttt{KMT2A/B/C/D}, \texttt{KMT5A/B}, \texttt{SUV39H1/2}, \texttt{SETD2/DB1}, \texttt{KDM1A/2A/B/4A/5A/B/C}); histone acetyltransferases and deacetylases (\texttt{KAT2A/B/5/6A/7/8}, \texttt{EP300}, \texttt{CREBBP}, \texttt{HDAC1--6/8}); nuclear envelope (\texttt{LMNA}); transcription machinery (\texttt{TAF1/2/4/6/7/10}, \texttt{POLR2A/B/C}, \texttt{TRRAP}, \texttt{MYC}); DNA-damage response (\texttt{BRCA1/2}, \texttt{ATM}, \texttt{ATRX}, \texttt{PARP1}, \texttt{TOP2A/B}, \texttt{PCNA}); Polycomb/trithorax (\texttt{SUZ12}, \texttt{EED}, \texttt{RING1}, \texttt{BMI1}, \texttt{JARID2}, \texttt{PHF1/8}, \texttt{CBX2/7/8}); NuRD/co-repressor complexes (\texttt{RBBP4/7}, \texttt{MTA1/2}, \texttt{SIN3A/B}, \texttt{NCOR1/2}); and key nuclear signalling (\texttt{BRD2/4}, \texttt{MEN1}, \texttt{RUVBL1/2}).
Hits: \texttt{CHD4, KMT2B, EP300, TAF1, POLR2A, PCNA, TRRAP, MYC} (8 of 100).

\paragraph{Iteration 2 --- Gene-family exploitation begins.}
The agent explicitly analyses all eight hits and derives twelve expansion directions: from \texttt{TAF1} and \texttt{POLR2A} it targets remaining TAF subunits (\texttt{TAF3/5/8/9B/11--15}), all remaining POLR2 subunits (\texttt{POLR2D--L}), Pol~II CTD kinases (\texttt{CDK7/8/9/12/13}, cyclins \texttt{CCNT1/2}, \texttt{CCNH}), Mediator subunits (\texttt{MED1/4/12/14/17/23}), general transcription factors (\texttt{TBP}, \texttt{GTF2B/E1/F1/H1}, \texttt{YY1}, \texttt{SP1/3}), and elongation factors (\texttt{SUPT6H}, \texttt{SUPT16H}, \texttt{SNW1}, \texttt{TCERG1}); from \texttt{TRRAP} and \texttt{EP300} it adds TRRAP-associated HAT complexes (\texttt{JADE1/2/3}, \texttt{BRD3/8}); from \texttt{MYC} it tests \texttt{MYCN}, \texttt{MAX}, and \texttt{MXD1}; from \texttt{CHD4} it expands to \texttt{CHD5/6/9/1L}; from \texttt{KMT2B} it adds \texttt{KMT2E}, \texttt{KMT5C}, \texttt{DOT1L}, \texttt{SETD7}, \texttt{SMYD2/3}, \texttt{EHMT1/2}; and from \texttt{PCNA} it tests replication-fork partners (\texttt{RFC1--5}, \texttt{POLE}, \texttt{POLD1}).
Hits: \texttt{TAF5, SUPT6H, POLR3B} (3 of 100).

\paragraph{Iterations 3--10 --- Sustained exploitation with adaptive pivots.}
At the start of every iteration the agent re-reads all accumulated hits and derives the next batch by explicit pattern analysis, so its strategy genuinely evolves with feedback.
Iterations 3--5 continue systematic transcription-machinery sweeps seeded in iteration~2: Pol~III subunit expansion yields \texttt{POLR3K} (iter.~3) alongside Mediator hits \texttt{MED21/27}; nuclear-export and TFIIIC probes return \texttt{XPO1} and \texttt{GTF3C5} (iter.~4); DNA-replication/repair exploration adds \texttt{DNA2} and \texttt{UHRF1} (iter.~5).
Iteration~6 marks an inflection: the \texttt{TRIM28} and \texttt{MDM2} hits signal ubiquitin-pathway involvement, and the agent pivots accordingly in iteration~7 to the CUL3/RBX1/UBE2M E3-ligase complex and the PSMC2 proteasome subunit, yielding 6 hits in a single batch.
Iterations~8--9 exploit newly identified family anchors --- ZSCAN zinc-fingers (\texttt{ZSCAN9}), KLK serine proteases (\texttt{KLK7/10/11}), and PPP phosphatases (\texttt{PPP2CA}, \texttt{PPP1CB}, \texttt{PPP2R1A}) --- while iteration~10 returns no hits as the productive threads are exhausted. More precisely, KLK4 and ZSCAN5A were caught at iteration 6 and tagged at "diverse" in the LLM completion. The exploitation of those families begins at iteration 8 perhaps because other previously exploited families are exhausted as show in Table~\ref{tab:family_exploitation}.

\begin{table}[t]
\centering
\caption{Family exploitation and exhaustion across iterations for the ICL-EF agent (F0). Each cell shows hits/tested for that gene family at that iteration. Dashes indicate the family was not tested. Major early families (CHD, TAF, POLR, MED) are exhausted by iteration~7, forcing the agent to pivot to newly identified anchors (ZSCAN, KLK, PPP) at iterations~8--10. The table reports 26 hits that can be assigned to a gene family, 9 hits correspond to singleton genes (no family).}
\label{tab:family_exploitation}
\resizebox{.7\textwidth}{!}{%
\begin{tabular}{l c c c c c c c c c c c}
\toprule
Family & It.\,1 & It.\,2 & It.\,3 & It.\,4 & It.\,5 & It.\,6 & It.\,7 & It.\,8 & It.\,9 & It.\,10 & Total \\
\midrule
CHD & 1/5 & 0/4 & --- & --- & --- & --- & --- & --- & --- & --- & 1/9 \\
KMT2 & 1/4 & 0/1 & --- & --- & --- & --- & --- & --- & --- & --- & 1/5 \\
TAF & 1/6 & 1/14 & --- & 0/3 & --- & --- & --- & --- & --- & --- & 2/23 \\
POLR & 1/3 & 1/11 & 1/10 & --- & --- & --- & --- & --- & --- & --- & 3/24 \\
MED & --- & 0/6 & 2/10 & 0/1 & --- & --- & --- & --- & --- & --- & 2/17 \\
HDAC & 0/7 & --- & --- & --- & 0/4 & --- & --- & --- & --- & --- & 0/11 \\
SMARC/ARID & 0/10 & 0/2 & --- & 0/4 & --- & --- & --- & --- & --- & --- & 0/16 \\
KDM & 0/7 & --- & --- & 0/3 & --- & --- & --- & --- & --- & --- & 0/10 \\
CUL & --- & --- & --- & --- & --- & --- & 1/5 & 0/3 & --- & --- & 1/8 \\
TRIM & --- & --- & --- & --- & --- & 1/3 & --- & 0/8 & 0/3 & 0/9 & 1/23 \\
IRF & --- & --- & --- & --- & --- & --- & 1/7 & 0/2 & 0/1 & --- & 1/10 \\
CDK & --- & 0/5 & 0/4 & --- & --- & --- & --- & --- & --- & --- & 0/9 \\
PSM & --- & --- & --- & --- & --- & --- & 1/4 & 0/2 & 0/13 & 0/5 & 1/24 \\
UBE2 & --- & --- & --- & --- & --- & --- & 1/5 & --- & 0/4 & 0/19 & 1/28 \\
USP & --- & --- & --- & --- & --- & --- & 0/1 & 0/12 & --- & --- & 0/13 \\
ZSCAN & --- & --- & --- & --- & --- & 1/1 & --- & 1/12 & --- & --- & 2/13 \\
KLK & --- & --- & --- & --- & --- & 1/1 & --- & 3/10 & 0/3 & 0/1 & 4/15 \\
PPP & --- & --- & --- & --- & --- & --- & --- & 1/1 & 2/4 & 0/18 & 3/23 \\
KIF & --- & --- & --- & --- & --- & --- & --- & 1/4 & 0/7 & --- & 1/11 \\
GTF & --- & 0/4 & --- & 1/14 & 0/1 & --- & --- & --- & --- & --- & 1/19 \\
SUPT & --- & 1/2 & 0/4 & --- & --- & --- & --- & --- & --- & --- & 1/6 \\
MCM & --- & 0/2 & 0/5 & --- & --- & --- & --- & --- & --- & --- & 0/7 \\
RNF & --- & --- & --- & --- & --- & --- & 0/3 & --- & --- & 0/6 & 0/9 \\
SIRT & --- & --- & --- & --- & 0/3 & --- & --- & 0/3 & --- & --- & 0/6 \\
NUP & --- & --- & --- & 0/3 & --- & --- & --- & --- & --- & --- & 0/3 \\
FBX & --- & --- & --- & --- & --- & 0/2 & 0/1 & 0/6 & --- & 0/7 & 0/16 \\
PRMT & --- & --- & --- & --- & --- & --- & 0/4 & 0/2 & --- & --- & 0/6 \\
\bottomrule
\end{tabular}
}
\end{table}

\paragraph{Summary.}
After 1,000 tests the agent discovers 35 of 152 significant genes (23.0\% recall): \texttt{CHD4, KMT2B, EP300, TAF1, POLR2A, PCNA, TRRAP, MYC, TAF5, SUPT6H, POLR3B, POLR3K, MED21, MED27, XPO1, GTF3C5, DNA2, UHRF1, TRIM28, KLK4, ZSCAN5A, IRF3, MDM2, CUL3, RBX1, PSMC2, UBE2M, ZSCAN9, KLK7, KLK10, KLK11, KIF11, PPP2CA, PPP1CB, PPP2R1A}. 
The trajectory reveals a clear exploitation dynamic: transcription-machinery hits in iterations~1--5 trigger systematic family sweeps (TAF, POLR, Mediator, GTF subunits); iteration~6 seeds three non-obvious families via \texttt{TRIM28}, \texttt{KLK4}, and \texttt{ZSCAN5A}; iteration~7 then pivots to the ubiquitin-proteasome system, yielding 6 hits in a single batch (\texttt{IRF3}, \texttt{MDM2}, \texttt{CUL3}, \texttt{RBX1}, \texttt{PSMC2}, \texttt{UBE2M}); and iteration~8 expands the KLK and ZSCAN families while discovering PPP phosphatases (\texttt{PPP2CA}), which yield two further hits in iteration~9. This name-prefix-driven exploitation accounts for the performance gap over the zero-shot agent (see Appendix~\ref{app:simple}).


\section{ICBR-EF Agent: Prompting Strategy and Execution Trace}
\label{app:icbf}

The ICBR-EF (In-Context Belief Revision from Experimental Feedback) agent extends ICL-EF with two mechanisms. First, we provide a \emph{phenotypic fingerprint} that gives the LLM the top-10 co-affected CellProfiler features for a selection of the tested gene (8 most recent hits and 4 most recent misses), enabling mechanism-based clustering. 
Besides, we also compute co-occurring significant side-effects across all hits: for each successful gene, we count how many other CellProfiler features (besides the target) also reach p < 0.05. The top-5 most frequently co-significant features are then reported as a shared phenotypic signature, e.g. - AGP: 7/10 HITs.
This tells the LLM which secondary cellular changes reliably accompany the target effect yielding effectively a data-driven phenotypic cluster that can guide selection of untested genes likely to share the same mechanism.

Second, an explicit \emph{hypothesis register} (structured as a JSON) is updated by the LLM across iterations, tracking mechanistic beliefs with confidence levels (\texttt{High}/\texttt{Medium}/\texttt{Low}/\texttt{Neutral}) and statuses (\texttt{Active}/\texttt{Weakened}/\texttt{Abandoned}/\texttt{New}). Below we reproduce the exact system prompt, the hypothesis register schema, the user prompt structure, and an annotated execution trace.

\subsection{System Prompt}

The system prompt is fixed across all iterations. It is constructed at agent initialization by filling in the target feature name, pool size, and feature count:

\begin{lstlisting}
You are an expert computational biologist AI agent discovering gene perturbations
that induce significant cellular responses.

TASK: Find gene knockouts where p-value for 'Nuclei_Intensity_MeanIntensity_AGP'
      is < 0.05.

## STRATEGY: Phenotypic Clustering & Mechanism Inference

Your key advantage is seeing the FULL PHENOTYPIC FINGERPRINT of each tested gene:
- Target feature p-value (what we're optimizing)
- Top 10 side effect p-values (other affected features)

### Use phenotypes to identify mechanisms:
1. Phenotype Clustering: Genes with similar side-effect profiles likely share
   mechanisms
2. Pathway Inference: If a gene affects features X,Y,Z and is a HIT, test other
   genes that might affect the same features
3. Functional Categories: Group genes by phenotypic signature, not just name

### Hypothesis-Driven Selection:
- Form hypotheses about which mechanisms cause hits
- Test hypotheses by selecting genes predicted to share that mechanism
- Update confidence based on results

## AVAILABLE DATA:
- Total perturbations: 7975
- Features: 100 CellProfiler measurements
- Target: Nuclei_Intensity_MeanIntensity_AGP

## OUTPUT FORMAT:
Return a JSON object with:
1. "hypotheses_register": Updated list of your hypotheses with confidence levels
2. "selection": List of gene names to test next

Example:
{
  "hypotheses_register": [
    {
      "hypothesis": "Transporter family genes cause intensity changes",
      "confidence": "High",
      "status": "Active",
      "reasoning": "Multiple transporter genes showed strong effects"
    },
    {
      "hypothesis": "Kinases in general affect this phenotype",
      "confidence": "Low",
      "status": "Weakened",
      "reasoning": "Tested 3 kinases, all MISS"
    }
  ],
  "selection": ["GENE1", "GENE2", "GENE3", ...]
}

Remember: Use phenotypic similarity to guide selection, not just gene name patterns!
\end{lstlisting}

\subsection{Hypothesis Register Schema}

The hypothesis register is a JSON array, entirely authored and updated by the LLM. Each entry has four fields:

\begin{lstlisting}
[
  {
    "hypothesis": <string>,   // Free-text mechanistic claim
    "confidence": <string>,   // "High" | "Medium" | "Low" | "Neutral"
    "status":     <string>,   // "Active" | "Weakened" | "Abandoned" | "New"
    "reasoning":  <string>    // Evidence cited: gene names, p-values, phenotypes
  },
  ...
]
\end{lstlisting}

The framework seeds the register with a single entry before iteration~1:

\begin{lstlisting}
[{"hypothesis": "Global Exploration", "confidence": "Neutral", "status": "Active",
  "reasoning": "Starting broad search to identify active phenotypes."}]
\end{lstlisting}

At every subsequent iteration the LLM replaces this array entirely, adding, updating, or removing entries as evidence accumulates.

\subsection{Per-Iteration User Prompt}

At each iteration $t$, the user prompt is assembled dynamically from three components.

\paragraph{1. Observation summary.}  For each tested gene the agent receives a \emph{phenotypic fingerprint}: target p-value and the five most significant co-affected features (abbreviated CellProfiler names). For example, after the first batch of 100 genes:

\begin{lstlisting}
TESTED: 100 | HITs: 7 | MISSes: 93

RECENT HITs (with phenotypic fingerprints):
  EP300:  [HIT] Target_p=0.0073 | Side-effects: Cells_Texture_DifferenceEntropy_AGP=0.000, Cells_Texture_InverseDifferenceMoment_AGP=0.000, ...
  ...
  KMT2B:  [HIT] Target_p=0.0024 | Side-effects: Cells_Texture_DifferenceEntropy_AGP=0.000, Cells_Texture_InverseDifferenceMoment_AGP=0.000, ...

RECENT MISSes (sample):
  PRMT1:  [MISS] Target_p=0.7797 | Side-effects: Nuclei_Texture_AngularSecondMoment_AGP=0.543, Nuclei_Texture_AngularSecondMoment_AGP=0.551, ...
  ...
  MBD2:   [MISS] Target_p=0.4998 | Side-effects: Cells_Texture_InverseDifferenceMoment_AGP=0.597, Nuclei_Texture_InverseDifferenceMoment_AGP=0.597, ...
\end{lstlisting}

Once at least 3 hits have been observed, a co-occurrence analysis is appended to the summary. For each successful gene, all CellProfiler features reaching $p < 0.05$ (other than the target) are counted; the 5 most frequent are reported as a shared phenotypic signature:

\begin{lstlisting}
### PHENOTYPE PATTERNS IN HITs:

 Common side-effects in HITs: - Cytoplasm_Texture_Entropy_AGP: 6/7 HITs - Nuclei_Texture_AngularSecondMoment_AGP: 5/7 HITs - Cells_Texture_InverseDifferenceMoment_AGP: 5/7 HITs -  Nuclei_Texture_InverseDifferenceMoment_AGP: 5/7 HITs - Cytoplasm_Texture_InverseDifferenceMoment_AGP: 5/7 HITs 
\end{lstlisting}

This block tells the LLM which secondary phenotypic changes reliably co-occur with target-feature hits, providing a data-driven anchor for hypothesis formation and phenotypic clustering.

\paragraph{2. Current hypothesis register.} The full JSON array from the previous iteration is serialized and appended.
\begin{lstlisting}
### CURRENT HYPOTHESES:                                                                                                                                                                 
[ 
  { "hypothesis": "Chromatin remodeling and epigenetic regulators affect nuclear AGP intensity", 
    "confidence": "Medium", 
    "status": "Active", 
    "reasoning": "Histone modifiers (HDACs, EZH2, KDM1A, BRD4) alter chromatin compaction and nuclear architecture, potentially affecting AGP staining intensity in nuclei" 
  }, 
...
]  
\end{lstlisting}

\paragraph{3. Task specification.} We list already-tested and untested gene IDs, then asks the LLM to (a)~update the hypothesis register and (b)~propose the next 100 genes. An explicit self-check instruction is added:

\begin{lstlisting}
YOUR TASK:                                                                                                                                                                              
  1 Update your hypotheses based on the phenotypic patterns observed 
  2 Select 100 perturbations that will:                                                                                                                                                 
     - Test your high-confidence hypotheses (exploitation)
     - Explore new potential mechanisms (exploration)
     - Use phenotypic similarity to find related genes 


CRITICAL REQUIREMENT: Before returning your final JSON, you MUST explicitly double-check your proposed genes ONE BY ONE. For each gene you want to propose, verify:

  1 It is EXACTLY present in the UNTESTED list above.
  2 It is NOT present in the ALREADY TESTED list. If you realize a gene violates these rules, propose a replacement and double-check the replacement.     

After your verification, return the JSON strictly inside a markdown code block exactly like this:  
  {                                                                                                                                                                                     
    "hypotheses_register": [...],                                                                                                                                                       
    "selection": ["GENE1", "GENE2", ...]                                                                                                                                                
  } 
\end{lstlisting}

\subsection{Execution Trace}

We trace the agent on \texttt{Nuclei\_Intensity\_MeanIntensity\_AGP} / F0 (152 significant genes out of 7,975), with batch size 100 and 10 iterations (1,000 total tests, 12.5\% of the pool).

\paragraph{Iteration 1 --- Broad initialization.}
With no observations yet, the agent interprets the target feature as reflecting nuclear AGP-channel protein accumulation and constructs seven mechanism-grounded hypotheses: chromatin remodelling/epigenetics (HDACs, EZH2, BRD4), cell-cycle control (CDKs, Aurora kinases), nuclear transport (XPO1, importins, nucleoporins), DNA-damage response (ATM, ATR, BRCA1/2, PARP1), master transcription factors (MYC, SP1, CTCF), kinase signalling (AKT, MTOR, MAPK), and ubiquitin/proteasome (MDM2, CUL3).
It selects 100 genes spanning all seven categories.
Hits: EP300, XPO1, MYC, MDM2, CUL3, CHD4, KMT2B (7/100).

\paragraph{Iteration 2 --- Focused exploitation with hypothesis refinement.}
Observing that all 7 hits are major chromatin/ubiquitin/transcription regulators sharing AGP-texture side-effect signatures, the agent abandons the broad cell-cycle, DNA-damage-response, and kinase-signalling hypotheses (all returned only misses) and replaces them with eight targeted hypotheses: KMT-family histone methyltransferases (from KMT2B), CHD/SWI--SNF remodellers (from CHD4), NuRD complex subunits (CHD4 is a NuRD core component), cullin-RING ubiquitin ligases (from CUL3, the strongest hit), importin/exportin nuclear transport (from XPO1), MYC-network partners (MAX, MXI1, MXD1), Mediator subunits (from EP300/CREBBP), and HAT/coactivator complexes (from EP300).
The 100-gene batch covers all eight hypotheses.
Hits: UHRF1 (1/100).

\paragraph{Iteration 3 --- Pruning failed branches, opening new ones.}
The analysis of MISSes from iterations 1--2 leads the agent to drop four previously active hypotheses: HDAC-family members (HDAC1--8 all missed), CHD paralogs beyond CHD4 (CHD1/2/5/6/7/8/9 all missed), KMT paralogs beyond KMT2B (KMT2A/C/D/E all missed), and importin-side nuclear transport (most importins missed).
Four new hypotheses are raised: CUL3 substrate adaptors with BTB domains (KEAP1 and related, motivated by CUL3 being the strongest hit), PHD-finger chromatin readers (PHF family connecting to the NuRD/CHD4 hit), PRDM/SET-domain methyltransferases (extending from KMT2B), and transcription elongation factors (SUPT and TAF families, motivated by the EP300/MYC/TRRAP context).
The batch covers these four new strands alongside the continuing Mediator, NuRD, and cullin-RING hypotheses.
Hits: TRIM28, MED27, SUPT6H, TAF1, TAF5 (5/100).

\paragraph{Iterations 4--10 --- Cascading hypothesis expansion driven by new hits.}
Iterations 4--5 exploit the TAF/Mediator/SUPT leads from iteration~3, yielding \texttt{MED21}, \texttt{POLR2A}, \texttt{RBM14}, and \texttt{POLR3B}.
\texttt{POLR2A} triggers a major pivot: the agent raises a new high-confidence hypothesis on RNA Pol~II/III machinery and shifts selections towards RNA polymerase subunits, general transcription factors (\texttt{GTF} family), and transcription-associated kinases (\texttt{CDK7/8/9}).
Iteration~6 confirms this strand with five hits (\texttt{POLR3K}, \texttt{GTF3C5}, \texttt{PRPF19}, \texttt{EXOSC9}, \texttt{XAB2}), establishing RNA Pol~III transcription and pre-mRNA splicing/RNA surveillance as new dominant hypotheses.
Iteration~7 investigates DDX helicases and spliceosome components (which are co-transcriptional RNA processing extending the transcriptional strand), hitting \texttt{DDX39A}, \texttt{ZFP36L2}, \texttt{CAND1}, and \texttt{EIF3G}. The \texttt{ZFP36L2} hit (AU-rich element mRNA decay) opens yet another hypothesis strand on mRNA stability, while \texttt{EIF3G} suggests translation factors as a new target class.
Iterations 8--9 further expand into proteasome (\texttt{PSMC2}), DNA replication (\texttt{PCNA}, \texttt{DNA2}), RNA modification (\texttt{METTL1}), nuclear export (\texttt{RAE1}), ribosomal proteins (\texttt{RPL4}, \texttt{RPL7}), translation elongation (\texttt{EEF2}), and DNA repair (\texttt{PNKP}), accumulating 11 additional hits.
Iteration~10 returns zero hits as the agent exhausts high-confidence candidates and sweeps lower-priority targets (DNA polymerases, cohesin, checkpoint genes).
Table~\ref{tab:family_exploitation_icbref} report how gene family are exploited by the agent across iterations.

\paragraph{Summary.}
After 1,000 tests the agent discovers 37 of 152 significant genes (24.3\% recall): \texttt{EP300, XPO1, MYC, MDM2, CUL3, CHD4, KMT2B, UHRF1, TRIM28, TRRAP, MED27, SUPT6H, TAF1, TAF5, MED21, POLR2A, RBM14, POLR3B, POLR3K, PRPF19, GTF3C5, EXOSC9, XAB2, DDX39A, ZFP36L2, CAND1, EIF3G, PCNA, DNA2, POP4, METTL1, PSMC2, RAE1, EEF2, PNKP, RPL4, RPL7}.
The trajectory reveals a clear mechanistic drift: the first three iterations concentrate on chromatin regulators, ubiquitin ligases, and nuclear transport; iterations 4--6 pivot to the transcription machinery (TFIID, Mediator, RNA Pol II/III); iterations 7--9 cascade outward into RNA processing, mRNA stability, translation, ribosome biogenesis, proteasome, and DNA repair, each pivot triggered by a surprise hit whose phenotypic fingerprint points to an unexpected but related pathway.
The hypothesis register expands from 7 broad priors at iteration~1 to over 12 mechanistically distinct entries by iteration~9, spanning chromatin remodelling, ubiquitin/CUL3 ligases, Mediator/TFIID, RNA Pol~II/III, spliceosome, RNA surveillance/exosome, mRNA stability, translation, ribosome biogenesis, proteasome, DNA replication, and DNA repair that were all inferred autonomously from phenotypic co-occurrence patterns in the observed hits.

\begin{table}[t]
\centering
\caption{Family exploitation across iterations for the ICBR-EF agent (F0). Each cell shows hits/tested. The agent performs shallow within-family probing before rapidly pivoting across gene families that share high-level phenotypic signatures (e.g., RNA-binding, essentiality, central dogma involvement), despite limited functional or mechanistic continuity. The table reports 20 hits that can be assigned to a gene family, 17 hits correspond to singleton genes (no family).}
\label{tab:family_exploitation_icbref}
\resizebox{.7\textwidth}{!}{%
\begin{tabular}{l c c c c c c c c c c c}
\toprule
Family & It.\,1 & It.\,2 & It.\,3 & It.\,4 & It.\,5 & It.\,6 & It.\,7 & It.\,8 & It.\,9 & It.\,10 & Total \\
\midrule
CHD & 1/2 & 0/6 & --- & --- & --- & --- & --- & --- & --- & --- & 1/8 \\
KMT2 & 1/3 & 0/2 & --- & --- & --- & --- & --- & --- & --- & --- & 1/5 \\
TAF & --- & --- & 1/3 & 1/16 & 0/4 & --- & --- & --- & --- & --- & 2/23 \\
POLR & --- & --- & --- & --- & 1/12 & 2/12 & --- & --- & --- & --- & 3/24 \\
MED & --- & 0/4 & 1/9 & 1/4 & --- & --- & --- & --- & --- & --- & 2/17 \\
HDAC & 0/3 & 0/5 & --- & 0/3 & --- & --- & --- & --- & --- & --- & 0/11 \\
SMARC/ARID & 0/5 & 0/6 & --- & --- & --- & --- & --- & --- & --- & 0/4 & 0/15 \\
KDM & 0/6 & --- & 0/3 & --- & --- & --- & --- & --- & --- & --- & 0/9 \\
CUL & 1/3 & 0/4 & --- & --- & --- & --- & --- & --- & --- & 0/1 & 1/8 \\
TRIM & --- & --- & 1/3 & 0/32 & 0/8 & --- & --- & --- & --- & --- & 1/43 \\
CDK & 0/4 & --- & --- & --- & 0/6 & --- & --- & --- & --- & --- & 0/10 \\
PSM & 0/2 & --- & --- & --- & --- & --- & --- & --- & 1/20 & 0/2 & 1/24 \\
UBE2 & --- & --- & --- & --- & --- & --- & --- & --- & 0/4 & --- & 0/4 \\
GTF & --- & --- & --- & --- & 0/15 & 1/4 & --- & --- & --- & --- & 1/19 \\
SUPT & --- & --- & 1/5 & 0/1 & --- & --- & --- & --- & --- & --- & 1/6 \\
EXOSC & --- & --- & --- & --- & --- & 1/5 & --- & --- & --- & --- & 1/5 \\
PRPF & --- & --- & --- & --- & --- & 1/4 & --- & --- & --- & --- & 1/4 \\
DDX & 0/1 & --- & --- & --- & --- & 0/8 & 1/12 & 0/11 & 0/4 & --- & 1/36 \\
EIF & --- & --- & --- & --- & --- & 0/5 & 0/3 & 1/8 & 0/7 & 0/5 & 1/28 \\
METTL & --- & --- & --- & --- & --- & --- & --- & 1/6 & --- & --- & 1/6 \\
EEF & --- & --- & --- & --- & --- & --- & --- & --- & 1/7 & --- & 1/7 \\
MCM & --- & --- & --- & --- & --- & --- & --- & 0/7 & --- & --- & 0/7 \\
RNF & --- & 0/2 & 0/3 & 0/1 & --- & --- & --- & --- & --- & --- & 0/6 \\
SIRT & 0/3 & --- & --- & --- & --- & --- & --- & --- & --- & --- & 0/3 \\
NUP & 0/3 & --- & --- & --- & --- & --- & 0/1 & --- & --- & --- & 0/4 \\
PRMT & 0/2 & 0/3 & 0/1 & --- & --- & --- & --- & --- & --- & --- & 0/6 \\
\bottomrule
\end{tabular}%
}
\end{table}

\section{Additional statistical significance assessment of performance gaps}
\label{app:per-feat-stats}
In this appendix, we report pairwise per-feature significance of performance gaps in Table~\ref{tab:per-feat-p-values}. 
Those p-values are obtained by applying paired two-sided sign-flip permutation tests to the replicate-level performance differences within each feature.

There is strong evidence that all baselines and the investigated approaches outperform choosing random genes at random. 
This is also true for Random FB which is indicative that the LLM is probably able to filter out the noisy feedback and leverage its sole prior knowledge. 
Moreover, when using Claude Sonnet 4.6, ICL approaches very consistently outperforms the zero-shot agent. 
The significance of the comparison of ICL methods vs Random FB is less sharp but remains the observed trend. 
Finally, the performance gain of ICBR-EF vs ICL-EF cannot be confirmed on a per-feature basis. 

To clarify this latter point, we also provide 99\% hierarchical bootstrap confidence intervals obtained by resampling features and, within each sampled feature, resampling replicates in Table~\ref{tab:cis}. These intervals provide an uncertainty estimate for the average feature-wise performance gap that complements the permutation-based significance tests reported in the main text.
Significant performances correspond to confidence intervals in which zero is excluded. 
The ICBR-EF vs ICL-EF remains borderline with the reported interval upper-bound being equal to zero but this analysis supports the conclusions drawn from Table~.\ref{tab:full} 
In this more holistic way of assessing statistical significance, the added of value of ICL compared to Random FB is clear.

\begin{table}[ht]
\centering
\resizebox{\textwidth}{!}{
\begin{tabular}{lrrrrrrrrrrr}
\toprule
Pair & F0 & F10 & F20 & F30 & F40 & F50 & F60 & F70 & F80 & F90 & $p<0.01$ \\
\midrule
Random vs GP-UCB & \textbf{0.004} & \textbf{0.004} & \textbf{0.002} & \textbf{0.002} & \textbf{0.008} & 0.012 & \textbf{0.008} & \textbf{0.002} & \textbf{0.008} & \textbf{0.002} & 9/10 \\
Random vs Zero-shot (4.5) & \textbf{0.004} & 0.076 & \textbf{0.002} & \textbf{0.002} & \textbf{0.002} & \textbf{0.004} & 0.012 & \textbf{0.002} & \textbf{0.002} & \textbf{0.002} & 8/10 \\
Random vs ICL-EF (4.5) & \textbf{0.004} & \textbf{0.002} & \textbf{0.002} & \textbf{0.004} & \textbf{0.006} & \textbf{0.002} & \textbf{0.002} & \textbf{0.006} & \textbf{0.002} & \textbf{0.002} & 10/10 \\
Random vs Zero-shot (4.6) & \textbf{0.008} & \textbf{0.004} & \textbf{0.002} & \textbf{0.002} & \textbf{0.002} & \textbf{0.002} & \textbf{0.004} & \textbf{0.002} & \textbf{0.004} & \textbf{0.002} & 10/10 \\
Random vs Random FB & \textbf{0.008} & \textbf{0.006} & \textbf{0.006} & \textbf{0.002} & \textbf{0.004} & \textbf{0.002} & \textbf{0.010} & \textbf{0.006} & 0.074 & \textbf{0.010} & 9/10 \\
Random vs ICL-EF (4.6) & \textbf{0.002} & \textbf{0.002} & \textbf{0.002} & \textbf{0.002} & \textbf{0.002} & \textbf{0.002} & \textbf{0.002} & \textbf{0.002} & \textbf{0.002} & \textbf{0.002} & 10/10 \\
Random vs ICBR-EF (4.6) & \textbf{0.002} & \textbf{0.002} & \textbf{0.002} & \textbf{0.002} & \textbf{0.002} & \textbf{0.002} & \textbf{0.002} & \textbf{0.002} & \textbf{0.002} & \textbf{0.002} & 10/10 \\
GP-UCB vs Zero-shot (4.5) & 1.000 & 0.016 & 0.523 & 0.828 & 0.344 & 0.633 & 0.840 & 0.266 & 0.176 & 0.258 & 0/10 \\
GP-UCB vs ICL-EF (4.5) & 0.648 & 0.770 & 0.662 & 0.311 & 0.773 & 0.152 & 0.119 & 0.982 & \textbf{0.002} & 0.031 & 1/10 \\
GP-UCB vs Zero-shot (4.6) & 0.266 & 0.859 & 0.172 & 0.600 & 0.875 & 0.102 & 0.898 & 0.047 & 0.891 & 0.031 & 0/10 \\
GP-UCB vs Random FB & 0.553 & 0.223 & 0.051 & 0.275 & 0.555 & 0.293 & 1.000 & 0.117 & 0.057 & 0.035 & 0/10 \\
GP-UCB vs ICL-EF (4.6) & \textbf{0.002} & 0.021 & \textbf{0.002} & 0.021 & \textbf{0.004} & \textbf{0.008} & \textbf{0.008} & \textbf{0.002} & \textbf{0.002} & \textbf{0.004} & 8/10 \\
GP-UCB vs ICBR-EF (4.6) & 0.012 & \textbf{0.002} & \textbf{0.002} & \textbf{0.004} & \textbf{0.002} & \textbf{0.002} & 0.012 & \textbf{0.002} & \textbf{0.002} & \textbf{0.002} & 8/10 \\
Zero-shot (4.5) vs ICL-EF (4.5) & 0.500 & 0.020 & 0.928 & 0.031 & 0.941 & 0.344 & 0.244 & 0.410 & \textbf{0.008} & 0.312 & 1/10 \\
Zero-shot (4.5) vs Zero-shot (4.6) & 0.117 & 0.012 & 0.373 & 0.648 & 0.438 & 0.758 & 0.566 & 0.672 & 0.078 & 0.520 & 0/10 \\
Zero-shot (4.5) vs Random FB & 0.586 & 0.055 & \textbf{0.010} & 0.328 & 1.000 & 0.949 & 0.836 & 0.029 & \textbf{0.006} & \textbf{0.004} & 3/10 \\
Zero-shot (4.5) vs ICL-EF (4.6) & \textbf{0.008} & \textbf{0.002} & \textbf{0.002} & \textbf{0.002} & \textbf{0.002} & \textbf{0.002} & \textbf{0.006} & \textbf{0.002} & \textbf{0.002} & \textbf{0.004} & 10/10 \\
Zero-shot (4.5) vs ICBR-EF (4.6) & \textbf{0.002} & \textbf{0.002} & \textbf{0.004} & \textbf{0.004} & \textbf{0.002} & \textbf{0.010} & 0.021 & \textbf{0.002} & \textbf{0.002} & \textbf{0.002} & 9/10 \\
ICL-EF (4.5) vs Zero-shot (4.6) & 0.086 & 0.625 & 0.055 & 0.062 & 0.848 & 0.703 & 0.086 & 0.141 & \textbf{0.002} & 1.000 & 1/10 \\
ICL-EF (4.5) vs Random FB & 0.941 & 0.250 & 0.084 & 0.016 & 0.973 & 0.469 & 0.172 & 0.234 & \textbf{0.002} & \textbf{0.004} & 2/10 \\
ICL-EF (4.5) vs ICL-EF (4.6) & 0.062 & \textbf{0.008} & \textbf{0.002} & \textbf{0.002} & \textbf{0.010} & \textbf{0.004} & 0.441 & \textbf{0.002} & \textbf{0.002} & \textbf{0.008} & 8/10 \\
ICL-EF (4.5) vs ICBR-EF (4.6) & 0.053 & \textbf{0.002} & \textbf{0.004} & \textbf{0.002} & 0.016 & \textbf{0.006} & 0.312 & \textbf{0.002} & \textbf{0.006} & \textbf{0.002} & 7/10 \\
Zero-shot (4.6) vs Random FB & 0.082 & 0.344 & 0.033 & 0.734 & 0.477 & 0.703 & 0.832 & \textbf{0.008} & 0.102 & \textbf{0.008} & 2/10 \\
Zero-shot (4.6) vs ICL-EF (4.6) & \textbf{0.004} & 0.012 & \textbf{0.002} & \textbf{0.004} & \textbf{0.002} & 0.020 & \textbf{0.008} & \textbf{0.002} & \textbf{0.002} & 0.055 & 7/10 \\
Zero-shot (4.6) vs ICBR-EF (4.6) & \textbf{0.002} & \textbf{0.002} & \textbf{0.004} & \textbf{0.004} & \textbf{0.002} & \textbf{0.004} & \textbf{0.004} & \textbf{0.002} & \textbf{0.002} & \textbf{0.002} & 10/10 \\
Random FB vs ICL-EF (4.6) & 0.074 & 0.014 & \textbf{0.002} & 0.027 & \textbf{0.006} & 0.014 & 0.020 & \textbf{0.002} & \textbf{0.002} & \textbf{0.002} & 5/10 \\
Random FB vs ICBR-EF (4.6) & 0.117 & \textbf{0.002} & \textbf{0.002} & \textbf{0.004} & \textbf{0.008} & \textbf{0.004} & 0.025 & \textbf{0.002} & \textbf{0.002} & \textbf{0.004} & 8/10 \\
ICL-EF (4.6) vs ICBR-EF (4.6) & 0.740 & 0.016 & 0.836 & 0.184 & 0.572 & 0.512 & 0.672 & 0.121 & 0.646 & 0.012 & 0/10 \\
\bottomrule
\end{tabular}}
\caption{Per-feature paired sign-flip permutation p-values. Bold: $p<0.01$.\label{tab:per-feat-p-values}}
\end{table}

\begin{table}[ht]
\centering
\begin{tabular}{lrr}
\toprule
Pair & Obs.\ diff & 99\% CI \\
\midrule
\textbf{Random vs GP-UCB} & \textbf{-8.530} & \textbf{[-10.610,\ -6.580]} \\
\textbf{Random vs Zero-shot (4.5)} & \textbf{-9.080} & \textbf{[-11.540,\ -6.640]} \\
\textbf{Random vs ICL-EF (4.5)} & \textbf{-10.800} & \textbf{[-15.270,\ -7.800]} \\
\textbf{Random vs Zero-shot (4.6)} & \textbf{-9.390} & \textbf{[-12.000,\ -7.050]} \\
\textbf{Random vs Random FB} & \textbf{-7.360} & \textbf{[-9.650,\ -5.070]} \\
\textbf{Random vs ICL-EF (4.6)} & \textbf{-18.330} & \textbf{[-25.090,\ -12.610]} \\
\textbf{Random vs ICBR-EF (4.6)} & \textbf{-20.430} & \textbf{[-28.000,\ -13.900]} \\
GP-UCB vs Zero-shot (4.5) & -0.550 & [-2.760,\ +2.280] \\
GP-UCB vs ICL-EF (4.5) & -2.270 & [-6.630,\ +0.640] \\
GP-UCB vs Zero-shot (4.6) & -0.860 & [-2.700,\ +0.970] \\
GP-UCB vs Random FB & +1.170 & [-1.200,\ +3.680] \\
\textbf{GP-UCB vs ICL-EF (4.6)} & \textbf{-9.800} & \textbf{[-16.010,\ -5.230]} \\
\textbf{GP-UCB vs ICBR-EF (4.6)} & \textbf{-11.900} & \textbf{[-18.560,\ -6.530]} \\
Zero-shot (4.5) vs ICL-EF (4.5) & -1.720 & [-5.600,\ +1.390] \\
Zero-shot (4.5) vs Zero-shot (4.6) & -0.310 & [-2.940,\ +1.990] \\
Zero-shot (4.5) vs Random FB & +1.720 & [-1.490,\ +5.140] \\
\textbf{Zero-shot (4.5) vs ICL-EF (4.6)} & \textbf{-9.250} & \textbf{[-14.900,\ -4.690]} \\
\textbf{Zero-shot (4.5) vs ICBR-EF (4.6)} & \textbf{-11.350} & \textbf{[-18.580,\ -5.820]} \\
ICL-EF (4.5) vs Zero-shot (4.6) & +1.410 & [-2.030,\ +6.400] \\
ICL-EF (4.5) vs Random FB & +3.440 & [-0.630,\ +9.090] \\
\textbf{ICL-EF (4.5) vs ICL-EF (4.6)} & \textbf{-7.530} & \textbf{[-11.720,\ -3.660]} \\
\textbf{ICL-EF (4.5) vs ICBR-EF (4.6)} & \textbf{-9.630} & \textbf{[-15.030,\ -4.950]} \\
Zero-shot (4.6) vs Random FB & +2.030 & [-1.080,\ +5.170] \\
\textbf{Zero-shot (4.6) vs ICL-EF (4.6)} & \textbf{-8.940} & \textbf{[-15.330,\ -4.500]} \\
\textbf{Zero-shot (4.6) vs ICBR-EF (4.6)} & \textbf{-11.040} & \textbf{[-17.930,\ -5.810]} \\
\textbf{Random FB vs ICL-EF (4.6)} & \textbf{-10.970} & \textbf{[-18.550,\ -5.080]} \\
\textbf{Random FB vs ICBR-EF (4.6)} & \textbf{-13.070} & \textbf{[-21.080,\ -6.380]} \\
ICL-EF (4.6) vs ICBR-EF (4.6) & -2.100 & [-5.490,\ +0.000] \\
\bottomrule
\end{tabular}
\caption{Hierarchical bootstrap comparison (sorted by $p$-value). Bold rows: 99\% CI excludes 0.\label{tab:cis}}
\end{table}

\end{document}